\documentclass[review, authoryear]{elsarticle}

\usepackage{geometry}
\usepackage{graphicx}
\usepackage{diagbox}
\usepackage{subfigure}
\usepackage{multirow}
\usepackage{multicol}
\usepackage{amsmath,amsfonts,amssymb}
\usepackage{url}
\usepackage{csquotes}
\usepackage[ruled,vlined]{algorithm2e}
\usepackage[T1]{fontenc}
\usepackage{gensymb}
\usepackage[dvipsnames]{xcolor}
\usepackage{lscape}

\newcommand{\PreserveBackslash}[1]{\let\temp=\\#1\let\\=\temp}
\newcolumntype{C}[1]{>{\PreserveBackslash\centering}p{#1}}

\begin{document}

\begin{frontmatter}

\title{Multimodal sensor data fusion for in-situ classification of animal behavior using accelerometry and GNSS data}

\author[data61]{Reza~Arablouei\corref{cor1}}
\author[data61]{Ziwei~Wang}
\author[af]{Greg~J.~Bishop-Hurley}
\author[data61]{Jiajun~Liu}
\cortext[cor1]{Corresponding author}
\address[data61]{Data61, CSIRO, Pullenvale QLD 4069, Australia}
\address[af]{Agriculture and Food, CSIRO, St Lucia QLD 4067, Australia}

\begin{abstract}

In this paper, we examine the use of data from multiple sensing modes, i.e., accelerometry and global navigation satellite system (GNSS), for classifying animal behavior. We extract three new features from the GNSS data, namely, distance from water point, median speed, and median estimated horizontal position error. We combine the information available from the accelerometry and GNSS data via two approaches. The first approach is based on concatenating the features extracted from both sensor data and feeding the concatenated feature vector into a multi-layer perceptron (MLP) classifier. The second approach is based on fusing the posterior probabilities predicted by two MLP classifiers. The input to each classifier is the features extracted from the data of one sensing mode. We evaluate the performance of the developed multimodal animal behavior classification algorithms using two real-world datasets collected via smart cattle collar tags and ear tags. The leave-one-animal-out cross-validation results show that both approaches improve the classification performance appreciably compared with using data of only one sensing mode. This is more notable for the infrequent but important behaviors of walking and drinking. The algorithms developed based on both approaches require little computational and memory resources hence are suitable for implementation on embedded systems of our collar tags and ear tags. However, the multimodal animal behavior classification algorithm based on posterior probability fusion is preferable to the one based on feature concatenation as it delivers better classification accuracy, has less computational and memory complexity, is more robust to sensor data failure, and enjoys better modularity.

\end{abstract}

\begin{keyword}
animal behavior classification \sep accelerometer data \sep deep learning \sep embedded systems \sep global navigation satellite system \sep information fusion \sep multimodal sensing \sep on-device inference.
\end{keyword}

\end{frontmatter}

\section{Introduction}

Accurate knowledge of the behavior of livestock is important for monitoring and managing their health, welfare, and productivity. However, long-term direct observation of behavior is challenging, particularly for large numbers of animals spread over large areas. This has motivated extensive research on the use of wearable sensors for automated remote recognition of various animal behaviors. Micro-electromechanical systems (MEMS) inertial measurement unit (IMU) sensors, particularly accelerometers, are among the most popular sensors with small form factor and low power consumption that have been successfully utilized for classification of animal behavior. Examples include the works of~\citep{aa1, sp1, gf, ca12} and the references therein.

The global navigation satellite system (GNSS) receiver is another device that is commonly used as a wearable sensor on animals of various species and sizes, both livestock and wildlife, to localize and track their positions. There have been some reports in the literature on using the data produced by GNSS receivers alone to classify animal behavior, for example, in~\citep{gn_c4, gn_c1, gn_sp1, gn_sb2}. Most of these works, calculate features from GNSS data over time segments that are significantly longer than those typically used when extracting features from IMU data. For instance, in~\citep{gn_c1}, it is shown that using segments of 160 seconds results in better performance compared with using shorter time segments. Meanwhile, accelerometry features used for animal behavior classification are often extracted from the data of time segments smaller than 15 seconds~\citep{ca0}. This difference is partly due to the frequency of GNSS data acquisition being set to much lower values compared with the sampling rate of accelerometer sensors as GNSS receivers consume significantly more power. The highest rate in most commercial GNSS receivers is 1 Hz while the sampling rate for MEMS accelerometers can be up to several KHz with typical values used for animal behavior classification being around 10 to 50 Hz~\citep{ca0}. The amount of discriminative information in the data and the associated uncertainty are other key factors determining the time segment size. The features that are commonly extracted from GNSS data for animal behavior classification are based on the distance traveled by the animal and the rate of change in the heading measurements, which relate to the direction of animal's movement. In addition, the range of behaviors that can be classified using GNSS data is usually more limited compared with the behaviors that can be classified using accelerometry data.

Accelerometry and GNSS data can provide discriminative knowledge regarding animal behaviors of interest that complement each other. This is because they may carry distinct information about the behaviors while being governed and affected by statistically independent processes. Some researchers have explored the advantages of the combined use of accelerometry and GNSS data for animal behavior classification~\citep{g+a5, g+a4, g+a1, g+a2}. In~\citep{g+a1}, several features extracted from accelerometry, GNSS, and weather data are used together with a support vector machine classifier to recognize behaviors that are relevant to detecting parturition events in grazing ewes. In~\citep{g+a2}, the accelerometry data is used for classifying the behavior of individual cattle while the GNSS data is used to detect and track cattle herds and their dispersion. In~\citep{g+a4}, twenty accelerometry features and three GNSS features are collectively used to distinguish between two basic cattle behaviors of grazing and non-grazing.

In this work, we study making use of the complementary information available through the accelerometry and GNSS data to enhance the accuracy of animal behavior classification. We extract three new features from the GNSS data, which relate to distance to the water point, Doppler-effect-borne speed, and estimated horizontal position error. We develop two algorithms via two different approaches for utilizing both accelerometry and GNSS features to classify animal behavior. The first approach is based on the concatenation of the accelerometry and GNSS features and using a multilayer perceptron (MLP) classifier. The second approach is based on classifying animal behavior using two MLP classifiers, each taking the features of one sensing mode, i.e., accelerometry or GNSS features, as input followed by fusing the posterior class probabilities predicted by the two MLP classifiers and inferring the behavior class from the fused posterior probabilities. This second approach has been adopted in various machine learning applications involving the combination of multiple classifiers~\citep{cc}. However, to the best of our knowledge, it has not been considered for animal behavior classification using multimodal sensor data.

We use two datasets that contain concurrent accelerometry and GNSS data to evaluate the performance of the developed multimodal animal behavior classification algorithms. The datasets are collected via smart cattle collar tags and ear tags during a field trail. We make these datasets publicly available\footnote{\url{https://github.com/Reza219/Animal_behavior_classification_Acc_GNSS}}. The leave-one-animal-out cross-validation results show that the combined utilization of the accelerometry and GNSS data offers improved classification accuracy over using the data of only one sensing mode. In addition, we observe that the algorithm based on the posterior probability fusion approach is more accurate, efficient, and scalable compared with the one based on concatenating the features despite its underlying independence assumption. The devised algorithms can be extended to be used with different or higher number of sensing modes. The advantages of the approach based on posterior probability fusion are likely to be more pronounced when the number of sensing modes is high and the statistical interdependence of the data coming from different sensing modes given the target class is sufficiently low.

\section{Calculation of features}\label{feat}

In this section, we describe the features that we calculate from accelerometry and GNSS data.

\subsection{Features of accelerometry data}\label{acc_feat}

The features that we calculate from accelerometry data are similar to those proposed in~\citep{RA21}, i.e., the mean and mean-absolute-filtered features. We describe their calculation in the following.

Let the $N$-dimensional vectors $\mathbf{a}_{ix}$, $\mathbf{a}_{iy}$, and $\mathbf{a}_{iz}$ denote the triaxial accelerometer readings associated with the $i$th datapoint ($i=1,2,\dots$) in three spatial axes, i.e., $x$, $y$, and $z$, respectively. In this work, we consider $N=256$.

To calculate the mean features, we average the entries of $\mathbf{a}_id$ for each axis $d\in\{x,y,z\}$ as
\begin{equation}
    m_{id}=\frac{1}{N}\mathbf{1}^{\intercal}\mathbf{a}_{id},\ d\in\{x,y,z\}
\end{equation}
where $\mathbf{1}$ is the all-ones column vector of appropriate size. These features relate to the animal's neck/head pose.

To calculate the next set of features, we pass the accelerometer readings of each axis through a first-order high-pass Butterworth filter that has a single parameter $\gamma_d$, $d\in\{x,y,z\}$. These infinite impulse response (IIR) filters remove the effect of gravity. We express their implementation by
\begin{equation}
    \left[1,-\gamma_d\right]^{\intercal}*\tilde{\mathbf{a}}_{id}=\left[1,-1\right]^{\intercal}*\mathbf{a}_{id},\ d\in\{x,y,z\}
\end{equation}
where $\tilde{\mathbf{a}}_{id}$ is the filter output and $*$ denotes the linear convolution operation. This notation is a time-domain representation of the utilized IIR filters, which we implement recursively as described in~\citep{RA21}.

We average the absolute values of the high-pass-filtered accelerometer readings for each axis to calculate the second set of features as
\begin{equation}
    s_{id}=\frac{1}{N}\mathbf{1}^{\intercal}|\tilde{\mathbf{a}}_{id}|,\ d\in\{x,y,z\}.
\end{equation}
These features relate to the intensity of the animal's body movements.

We collect both sets of above-mentioned features, calculated from the accelerometry data of the datapoint $i$, in the feature vector $\mathbf{f}_{ia}\in\mathbb{R}^{6\times 1}$.

\subsection{Features of GNSS data}

We calculate three GNSS features for each datapoint indexed by $i$ using four sets of values reported by the GNSS receiver. These values are latitude, longitude, speed, and estimated horizontal position error (EHPE), which are provided by the GNSS receiver during the time segment corresponding to each datapoint. The GNSS receiver data is reported approximately every one second. Therefore, during a typical time segment of five seconds, on average, five sets of GNSS data are available. The EHPE is the radius of a circle around the actual location that is estimated to contain the GNSS-estimated position with the probability associated with one standard-deviation of uncertainty, which is approximately 68\% if we assume Gaussian distribution for the perturbation causing the uncertainty. Here, we simply refer to EHPE as \emph{error}.

We compute two features, denoted by $v_{i}$ and $e_{i}$, as the medians of the speed and error values, respectively. The median speed feature, $v_{i}$, can be linked to the approximate distance traveled by the animal during the time window of each datapoint. Since GNSS speed is calculated using Doppler shift, it provides a better estimate of traveled distance per unit time compared to the distance between consecutive positions reported by the GNSS receiver. We choose to compute the median values instead of the means to have features that are more robust to outliers and erroneous readings.

We obtain another feature by calculating the distance of the water point from the median position of the collar, whose coordinates are the medians of the corresponding latitude and longitude values for each datapoint. We keep the water point at a fixed location during every day of data collection experiments and record its coordinates daily. To calculate the distance to the water point (DtWP) feature, we use the Pythagorean theorem on an equirectangular projection. This method is approximate. However, it is sufficiently accurate for short distances while being more computationally efficient compared to more accurate methods such as the haversine formula. Hence, we compute the DtWP feature, denoted by $\delta_{i}$, as

\begin{align}
  \delta_{i} &= r\sqrt{\cos^2\left(\frac{\tilde{\phi}_i+\phi_w}{2}\right)\left(\tilde{\lambda_i}-\lambda_w\right)^2+\left(\tilde{\phi_i}-\phi_w\right)^2}
\end{align}
where $r$ is the radius of earth (approximately 6,371,230 meters), $\phi_w$ and $\lambda_w$ are, respectively, the known latitude and longitude of the water point, and $\tilde{\phi_i}$ and $\tilde{\lambda_i}$ are the median latitude and longitude values for the $i$th datapoint, respectively.

We stack the three features of speed, error, and DtWP calculated from the GNSS data of the datapoint $i$ in the feature vector $\mathbf{f}_{ig}\in\mathbb{R}^{3\times 1}$.

\section{Classification of behavior}\label{alg}

In this section, we describe two methods for classifying animal behavior through combining the information coming from accelerometry and GNSS data.

\subsection{Concatenation of features}

The first method that we consider is to concatenate the feature vectors corresponding to the accelerometry and GNSS data and classify the behavior using the augmented feature vector, i.e., $\mathbf{f}_i=\left[\mathbf{f}_{ia}^{\intercal},\mathbf{f}_{ig}^{\intercal}\right]^{\intercal}$.

For the classification part, we feed the feature vector $\mathbf{f}_{i}$ into an MLP classifier that has a single hidden layer and uses the rectified linear unit (ReLU) activation function. We express the output of the MLP as
\begin{equation}
    \mathbf{z}_i = \mathbf{W}_{2}\max\left(\mathbf{0},\mathbf{W}_{1}\mathbf{f}_{i}+\mathbf{b}_{1}\right)+\mathbf{b}_{2}
\end{equation}
where $\mathbf{0}$ is the zero vector, $\mathbf{W}_1\in\mathbb{R}^{L\times F}$ and $\mathbf{b}_1\in\mathbb{R}^{L\times 1}$ are the weight matrix and the bias vector of the hidden layer, $\mathbf{W}_2\in\mathbb{R}^{C\times L}$ and $\mathbf{b}_2\in\mathbb{R}^{C\times 1}$ are the weight matrix and the bias vector of the output layer, $F$ is the number of features, $L$ is the dimension of the output of the hidden layer, and $C$ is the number of classes.

We use the $\mathrm{softmax}$ function to convert the output of the MLP to the normalized scores, which represent the posterior probabilities of belonging to the considered behavior classes, given the model is well calibrated. Therefore, we calculate the posterior probabilities as
\begin{equation}
p(y_c|\mathbf{f}_{i}) = \frac{e^{z_{ic}}}{\mathbf{1}^\intercal e^{\mathbf{z}_i}},\ c=1,\dots,C  
\end{equation}
where $y_c$ is the binary variable that indicates association to class $c\in\{1,\dots,C\}$, $z_{ic}$ is the $c$th entry of $\mathbf{z}_i$, and the exponentiation in the denominator is applied element-wise.

At training time, we require the normalized scores or the posterior probabilities $p(y_c|\mathbf{f}_{i})$ to calculate the cross-entropy loss associated with datapoint $i$. However, at inference time, to save on computations, we determine the inferred behavior class by finding the largest corresponding MLP output without applying the $\mathrm{softmax}$ function, i.e., by calculating $\arg\max_c{z_{ic}}$.

\begin{figure}
    \centering
    \includegraphics[scale=.9]{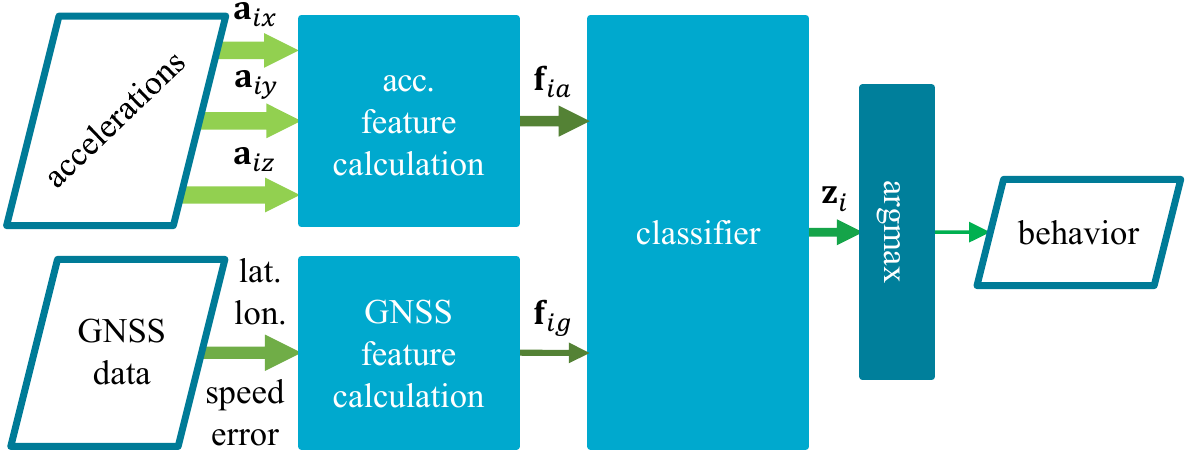}
    \caption{The block diagram of the animal behavior classification algorithm using multimodal sensor data via the feature concatenation method.}
    \label{conct}
\end{figure}

We depict the block diagram of the above-described method based on feature concatenation in Fig.~\ref{conct}.

\begin{figure}
    \centering
    \includegraphics[scale=.9]{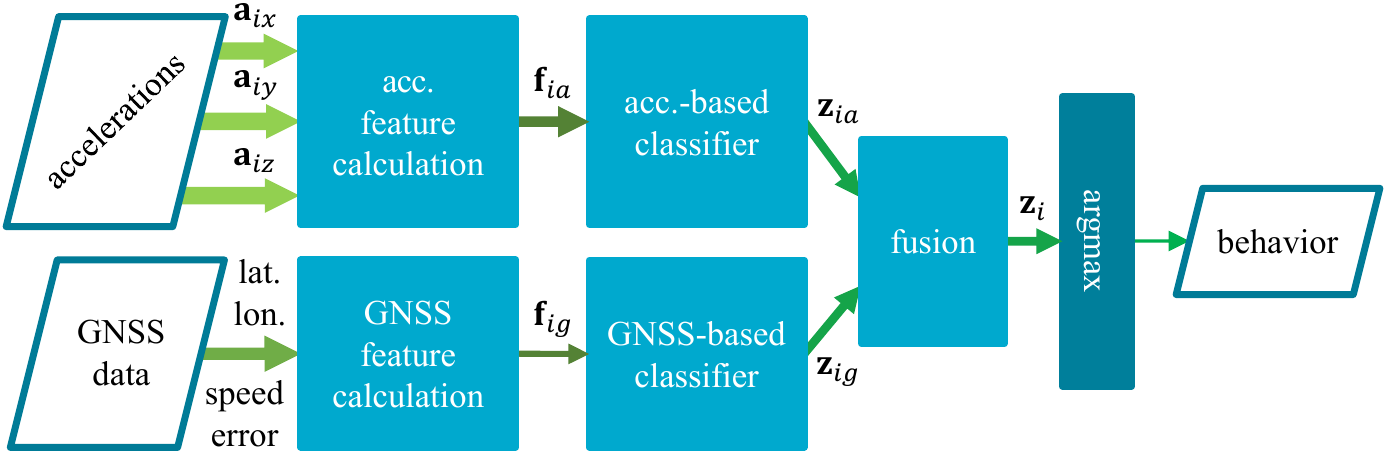}
    \caption{The block diagram of the animal behavior classification algorithm using multimodal sensor data via the posterior probability fusion method.}
    \label{fuse}
\end{figure}

\subsection{Fusion of posterior probabilities}

The second method that we consider is based on fusing the posterior probabilities predicted by two MLP classifiers that are separately fed with the features calculated from the accelerometry and GNSS data as shown in Fig.~\ref{fuse}. Each MLP classifier has one hidden layer and both MLPs use the ReLU activation function. The outputs of the MLPs corresponding to the accelerometry and GNSS data, respectively, are
\begin{align}
    \mathbf{z}_{ia} &= \mathbf{W}_{2a}\max\left(\mathbf{0},\mathbf{W}_{1a}\mathbf{f}_{ia}+\mathbf{b}_{1a}\right)+\mathbf{b}_{2a}\\
    \mathbf{z}_{ig} &= \mathbf{W}_{2g}\max\left(\mathbf{0},\mathbf{W}_{1g}\mathbf{f}_{ig}+\mathbf{b}_{1a}\right)+\mathbf{b}_{2g}
\end{align}
where $\mathbf{W}_{1a}$, $\mathbf{W}_{2a}$, $\mathbf{W}_{1g}$, and $\mathbf{W}_{2g}$ are the relevant weight matrices and $\mathbf{b}_{1a}$, $\mathbf{b}_{2a}$, $\mathbf{b}_{1g}$, and $\mathbf{b}_{2g}$ are the related bias vectors. Applying the $\mathrm{softmax}$ function to the MLP outputs yields the corresponding posterior probabilities as
\begin{align}
    p(y_c|\mathbf{f}_{ia}) &= \frac{e^{z_{iac}}}{\mathbf{1}^\intercal e^{\mathbf{z}_{ia}}},\ c=1,\dots,C \label{sma} \\
    p(y_c|\mathbf{f}_{ig}) &= \frac{e^{z_{igc}}}{\mathbf{1}^\intercal e^{\mathbf{z}_{ig}}},\ c=1,\dots,C \label{smg} 
\end{align}
where $z_{iac}$ and $z_{igc}$ are the $c$th entries of $\mathbf{z}_{ia}$ and $\mathbf{z}_{ig}$, respectively.

Here, our objective is to compute the posterior probability for each behavior class given the feature vectors $\mathbf{f}_{ia}$ and $\mathbf{f}_{ig}$, i.e., $p(y_c|\mathbf{f}_i)$, $c\in\{1,\dots,C\}$, using the sensor-specific posterior probabilities, i.e., $p(y_c|\mathbf{f}_{ia})$ and $p(y_c|\mathbf{f}_{ig})$, $c\in\{1,\dots,C\}$, produced by the respective classifiers and the prior probability of each behavior class, i.e., $p(y_c)$, $c\in\{1,\dots,C\}$.

\emph{Assumption}: To make the fusion of the posterior probabilities tractable, we assume that the likelihoods of observing $\mathbf{f}_{ia}$ and $\mathbf{f}_{ig}$ are conditionally independent of each other given the behavior class $y_c$ for every class and datapoint, i.e., $p\left(\mathbf{f}_{ia}|y_c\right)$ and $p\left(\mathbf{f}_{ig}|y_c\right)$ $\forall i \mathrm{\ and\ } c\in\{1,\dots,C\}$ are statistically independent. Therefore, we have
\begin{equation}
p\left(\mathbf{f}_{i}| y_{c}\right)=p\left(\mathbf{f}_{ia}| y_c\right)p\left(\mathbf{f}_{ig}| y_c\right),\ \forall i \mathrm{\ and\ } c\in\{1,\dots,C\}.
\end{equation}
We also make the implicit assumption that the behavior classes are mutually exclusive that means only one behavior class can be associated with each feature vector.

Using the Bayes theorem and the above independence assumption, we have
\begin{align}
p\left(y_c|\mathbf{f}_{i}\right)&=\frac{p\left(\mathbf{f}_i|y_c\right)p\left(y_c\right)}{p\left(\mathbf{f}_i\right)}\\
&=\frac{p\left(\mathbf{f}_{ia}|y_c\right)p\left(\mathbf{f}_{ig}|y_c\right)p\left(y_c\right)}{p\left(\mathbf{f}_i\right)}\\
&=\frac{p\left(\mathbf{f}_{ia}|y_c\right)p\left(y\right)p\left(\mathbf{f}_{ig}|y_c\right)p\left(y\right)}{p\left(\mathbf{f}_i\right)p\left(y\right)}\\
&=\frac{p\left(\mathbf{f}_{ia}\right)p\left(\mathbf{f}_{ig}\right)}{p\left(\mathbf{f}_i\right)}\times\frac{p\left(y_c|\mathbf{f}_{ia}\right)p\left(y_c|\mathbf{f}_{ig}\right)}{p\left(y_c\right)}\label{fus}\\
&\propto\frac{p\left(y_c|\mathbf{f}_{ia}\right)p\left(y_c|\mathbf{f}_{ig}\right)}{p\left(y_c\right)} \label{pro}
\end{align}
where $\propto$ denotes proportionality. The first multiplicand in~\eqref{fus} is independent of $y_c$ while the values of the features are known. Therefore, its value is the same for all classes making it irrelevant to classification. Consequently, to find the class with the highest fused posterior probability, we only need to find the class with the highest value of the right-hand side of~\eqref{pro}. Nonetheless, we can calculate the fused posterior probabilities for each class without the knowledge of the statistical distribution of the features via
\begin{align}
q_c\left(\mathbf{f}_{i}\right)&=\frac{p\left(y_c|\mathbf{f}_{ia}\right)p\left(y_c|\mathbf{f}_{ig}\right)}{p\left(y_c\right)}\\
p\left(y_c|\mathbf{f}_{i}\right)&=\frac{q_c\left(\mathbf{f}_{i}\right)}{\sum_{c'=1}^C{q_{c'}\left(\mathbf{f}_{i}\right)}},\ c\in\{1,\dots,C\}.
\end{align}

Taking the logarithm of both sides of~\eqref{pro} and using~\eqref{sma} and~\eqref{smg}, we have
\begin{align}
    \ln p\left(y_c|\mathbf{f}_{i}\right)
    &\propto\ln\frac{e^{z_{iac}}e^{z_{igc}}}{p\left(y_c\right)\mathbf{1}^\intercal e^{\mathbf{z}_{ia}}\mathbf{1}^\intercal e^{\mathbf{z}_{ig}}}\\
    &\propto z_{iac}+z_{igc}-\ln{p(y_c)}-\ln\left(\mathbf{1}^\intercal e^{\mathbf{z}_{ia}}\mathbf{1}^\intercal e^{\mathbf{z}_{ig}}\right)\\
    &\propto z_{iac}+z_{igc}-\ln{p(y_c)}.
\end{align}
Thus, we have
\begin{equation}
\arg \max_c{p\left(y_c|\mathbf{f}_{i}\right)} = \arg \max_c{\left(z_{iac}+z_{igc}-\ln{p(y_c)}\right)},
\end{equation}
which means, to determine the class with the highest fused posterior probability efficiently, we can calculate only the values of $z_{iac}+z_{igc}-\ln{p(y_c)}$, $c\in\{1,\dots,C\}$, and find the class with the largest corresponding value.

\section{Datasets}\label{datasets}

In this section, we describe two datasets containing both accelerometry and GNSS data that we use for evaluating the proposed multimodal animal behavior classification algorithms. We also explain the associated data collection experiment and annotation procedure.

\subsection{Data collection}

To evaluate the algorithms described in sections~\ref{feat} and~\ref{alg}, we use two datasets. We have obtained these datasets by running a field experiment for eight days during 18 to 27 March 2020 at the Commonwealth Scientific and Industrial Research Organisation (CSIRO) FD McMaster Laboratory Pasture Intake Facility~\citep{ts_armidale}, Chiswick NSW, Australia ($30\degree36'28.17''$S, $151\degree32'39.12''$E). The experiment was approved by the CSIRO FD McMaster Laboratory Chiswick Animal Ethics Committee with the animal research authority number 19/18.

We collected tri-axial accelerometry and GNSS data from eight Angus heifers using eGrazor collar tags\footnote{\url{https://www.csiro.au/en/research/animals/livestock/egrazor-measuring-cattle-pasture-intake}} and Ceres Tag ear tags\footnote{\url{https://www.cerestag.com/}}. All eight cattle were fitted with collar tags and six also wore ear tags. The experiment paddock consisted of four plots of approximately 25~x 100 m that were fully enclosed with electric fences. During each day of the experiment, the cattle were confined to an area of approximately 25 x 25 m. This way, the animals were in the field of view of the GoPro Hero 5 cameras\footnote{\url{https://gopro.com/}}, which we placed at four corners of the experiment area and used to record the behavior of the animals during the experiment with a resolution of $1920\times1080$ pixels and 30 frames per second.

Each eGrazor collar tag houses a 32GB SD flash memory card. However, the on-board flash memory of the Ceres Tag ear tags is limited to 1MB. Therefore, during the experiment, we logged the accelerometry and GNSS data of both collar tags and ear tags on the SD cards of the collar tags. To this end, we streamed the accelerometry and GNSS data of each ear tag to the collar tag worn by the same animal via a generic attribute profile (GATT) Bluetooth link. At the end of the experiment, we collected the SD cards and retrieved the logged data.

\begin{figure}
    \centering
    \includegraphics[scale=.72]{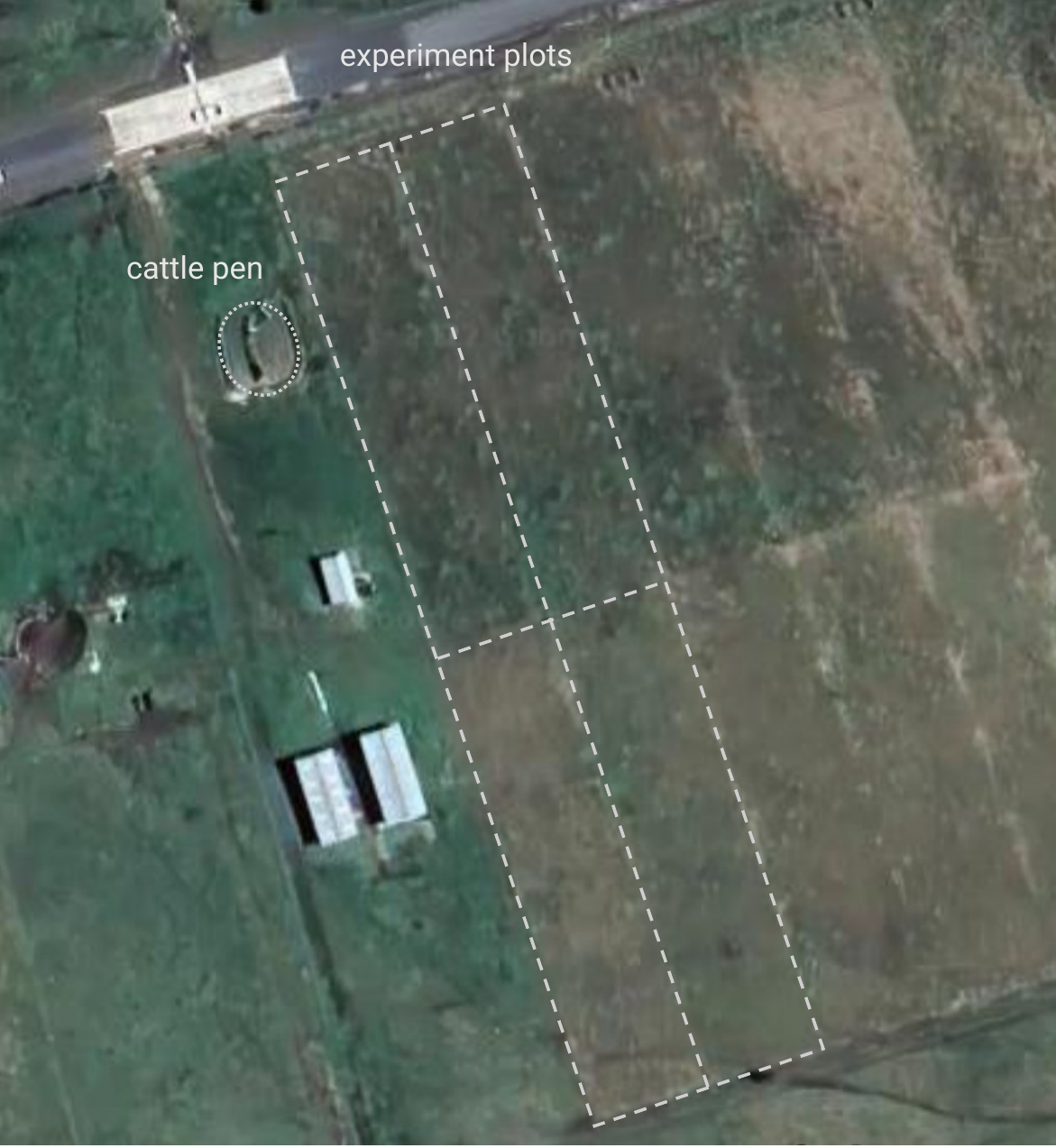}
    \caption{\small{The satellite view of the experiment site including the fence lines of the utilized plots and the location of the pen where cattle were kept at night.}}
   \label{sat}
\end{figure}

\begin{figure}
    \centering
    \includegraphics[scale=.35]{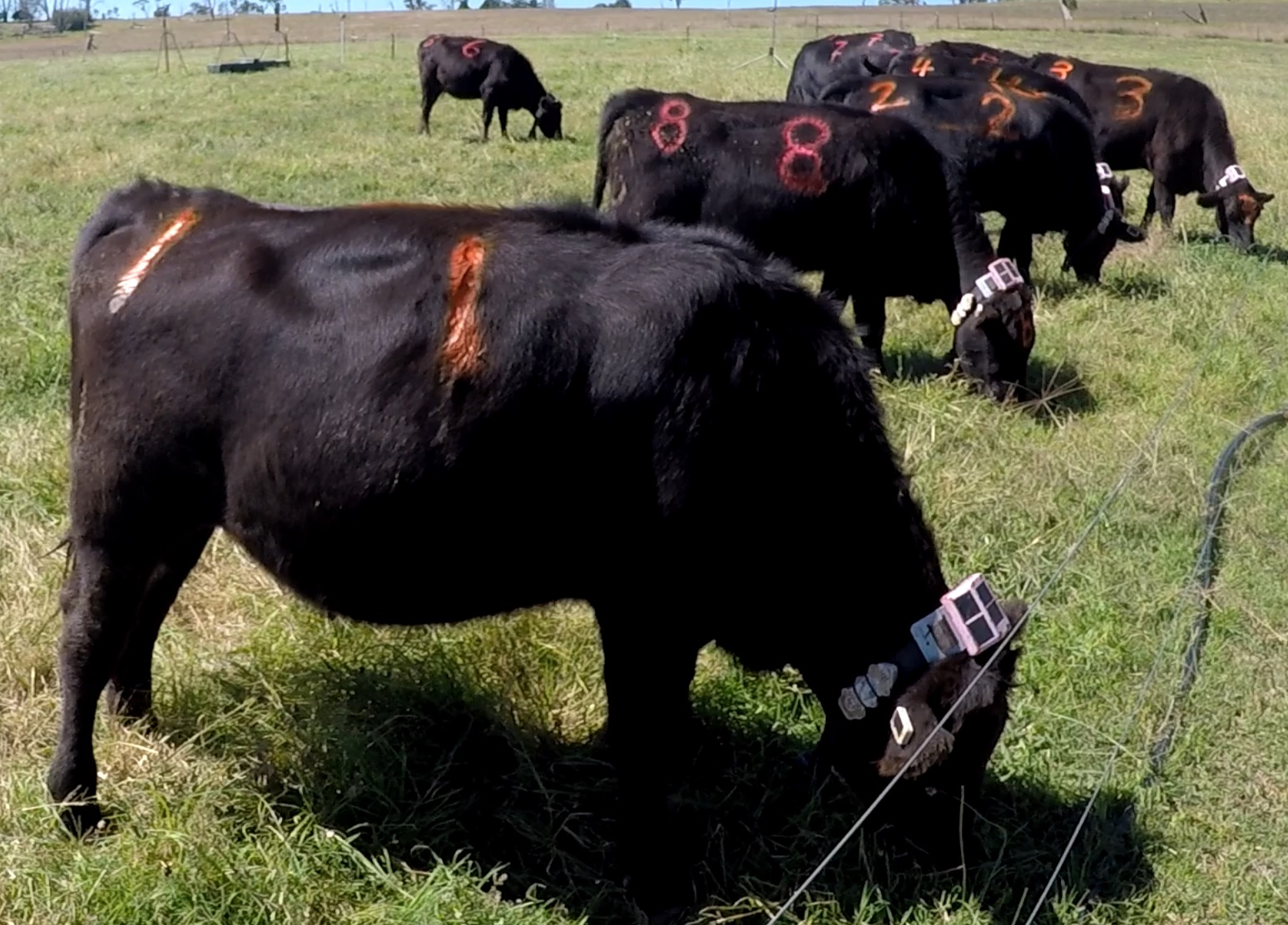}
    \caption{\small{Cattle on the paddock wearing collar and ear tags during the data collection experiment.}}
   \label{pad}
\end{figure}

Fig.~\ref{sat} is a satellite view of the experiment site where we have marked the fence lines of the utilized plots with dashed lines. The figure also includes the location of the pen that we used for installing and removing the collar and ear tags as well as keeping the cattle at night. Fig.~\ref{pad} is a snapshot of the paddock and the cattle used in the experiment wearing collar and ear tags. An overview of the hardware specifications for the eGrazor collar tag and Ceres Tag ear tag can be found in~\citep{dles} and~\citep{rnn_ab}, respectively.

The IMU used in the collar tags is InvenSense MPU-9250\footnote{\url{https://invensense.tdk.com/products/motion-tracking/9-axis/mpu-9250/}} and the accelerometer used in the ear tags is Bosch Sensortec BMA400\footnote{\url{https://www.bosch-sensortec.com/products/motion-sensors/accelerometers/bma400/}}. We set the accelerometer sample rate to $50$ Hz on the collar tags and to $62.5$ Hz on the ear tags. Therefore, the time window corresponding to $256$ consecutive accelerometer samples is about $5.12$ s for the collar tag accelerometry data and about $4.1$ s for the ear tag accelerometry data.
 
The GNSS receiver used in the collar tags is u-blox MAX-M8C\footnote{\url{https://www.u-blox.com/en/product/max-m8-series}} and the one used in the ear tags is u-blox ZOE-M8\footnote{\url{https://www.u-blox.com/en/product/zoe-m8-series}}. We logged the data reported by the GNSS receivers of the collar tags with the frequency of approximately once per second. Every ear tag also features a GNSS receiver. However, it is intended to be used with much lower frequency due to the small capacity of the ear tag's battery and photovoltaic module (solar panel). Therefore, the GNSS data that we use in this work came from the collar tags.

\subsection{Annotations}

We annotated sections of the collected accelerometry and GNSS data by observing the behavior of the cattle in the videos recorded via the GoPro cameras during the experiment. Experts logged the annotation instances using a mobile application specifically built for this purpose. We used the annotations to generate our labeled datasets called Arm20c and Arm20e, which correspond to the collar and ear tag data, respectively.

We consider five mutually-exclusive cattle behaviors of grazing, walking, resting, drinking, and alia. The resting behavior class also includes the ruminating behavior. The alia class encompasses all behaviors other than grazing, walking, resting, and drinking. To avoid confusion, we use the Latin word \textit{alia} instead of \textit{other}, which we have previously used for the same purpose.

We first use the annotations to label the corresponding accelerometry data. Afterwards, we create the datapoints of each dataset by dividing the related labeled tri-axial accelerometry data into non-overlapping segments of $256$ consecutive readings. We then find the GNSS receiver data whose timestamps fall into the time window of every segment in each dataset and associate them with the corresponding labeled accelerometry data to form the datapoints containing both accelerometry and GNSS data. We ignore the instances when either the accelerometry or GNSS data is missing. The median number of the GNSS readings of the datapoints is five in the Arm20c dataset and four in the Arm20e dataset. In Table~\ref{dataset}, we give the number of labeled datapoints for each behavior class in each dataset.

\begin{table} \footnotesize
\caption{The number of labeled datapoints for each behavior class in the considered datasets.} 
\vspace{12pt}
\label{dataset}
\centering
\begin{tabular}{|l| c c|}
\hline
\backslashbox{behavior}{dataset} & Arm20c & Arm20e\\
\hline
grazing  & 6156  & 5833 \\
walking  & 910   & 854  \\
resting  & 4080  & 3491 \\
drinking & 594   & 501  \\
alia    & 222   & 200  \\
\hline
total    & 11962 & 10879\\
\hline
\end{tabular}
\end{table}

\section{Results}

In this section, we evaluate the performance of the proposed multimodal behavior classification algorithms in terms of classification accuracy and computational complexity using the datasets described in section~\ref{datasets}. We also provide some visual analysis of the labeled GNSS data as well as the feature spaces associated with both accelerometry and GNSS data.

We make the utilized datasets as well as the Python code for reproducing the presented results publicly available\footnote{\url{https://github.com/Reza219/Animal_behavior_classification_Acc_GNSS}}.

\subsection{Visualization of GNSS data}\label{gnss_anal}

\begin{figure}
    \begin{multicols}{2}
    \centering
    \subfigure[23/March/2020]{\includegraphics[scale=.545]{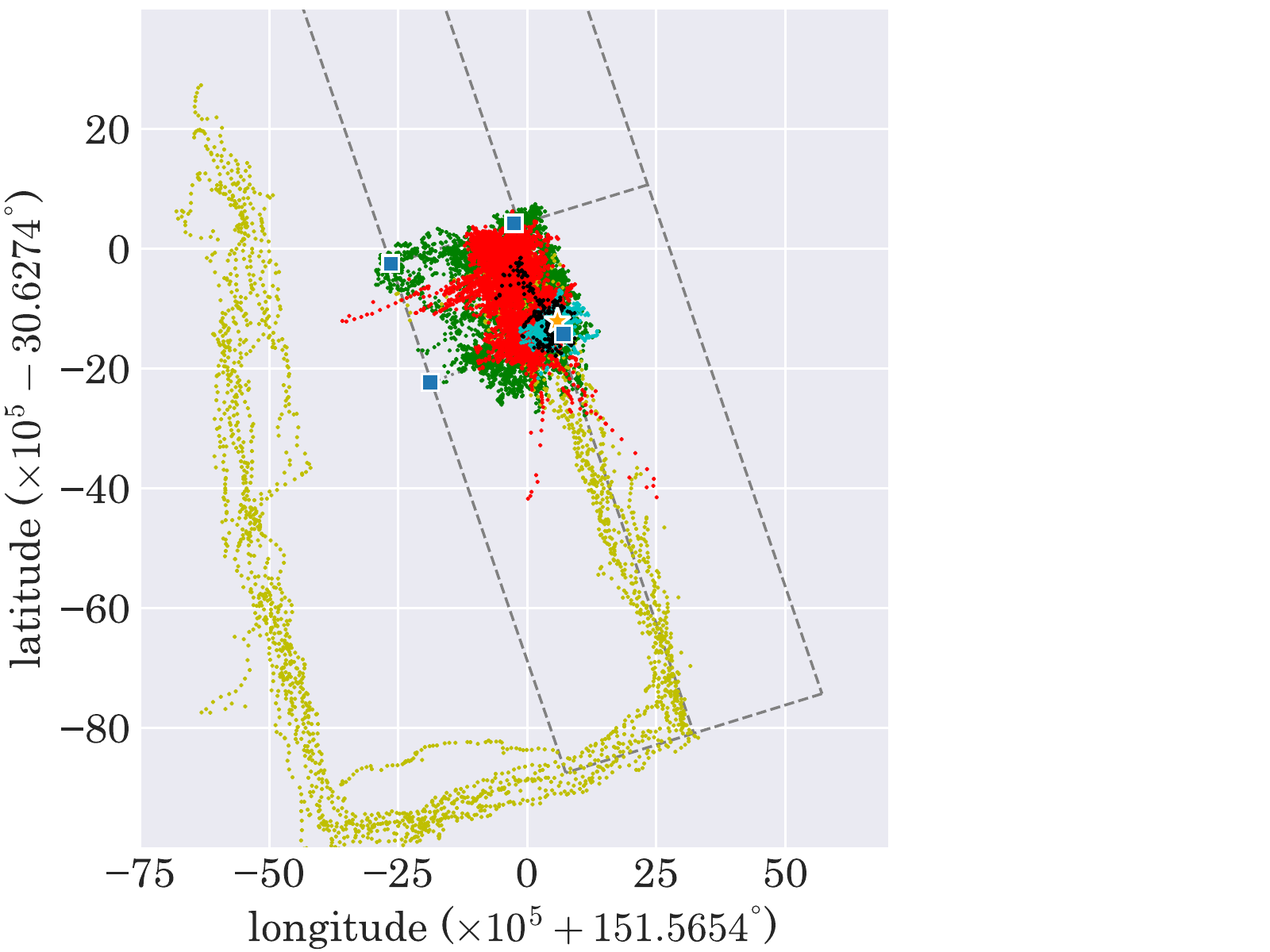}\label{23}}
    \subfigure[24/March/2020]{\includegraphics[scale=.545]{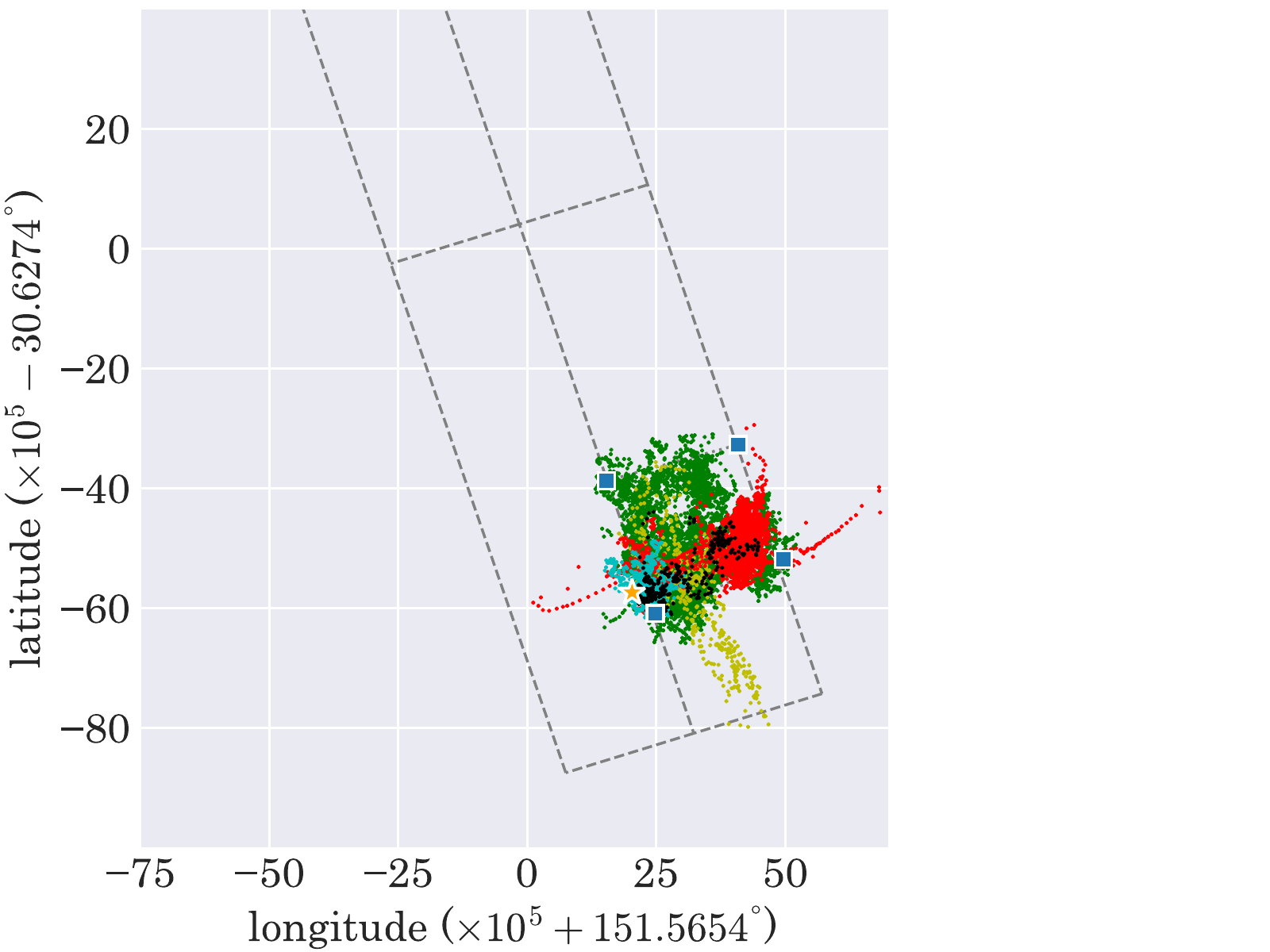}\label{24}}
    \subfigure[25/March/2020]{\includegraphics[scale=.545]{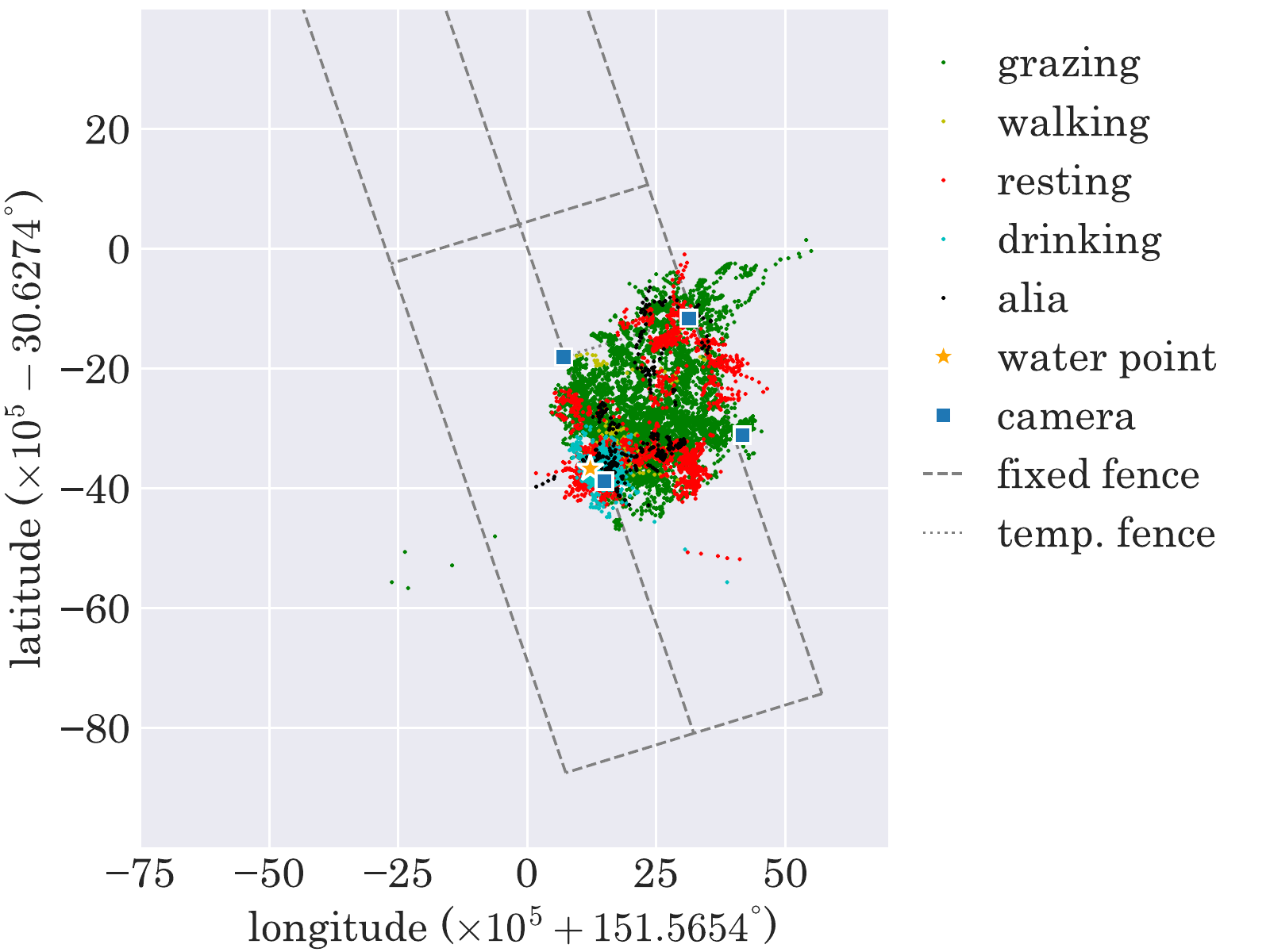}\label{25}}
    \subfigure[26/March/2020]{\includegraphics[scale=.545]{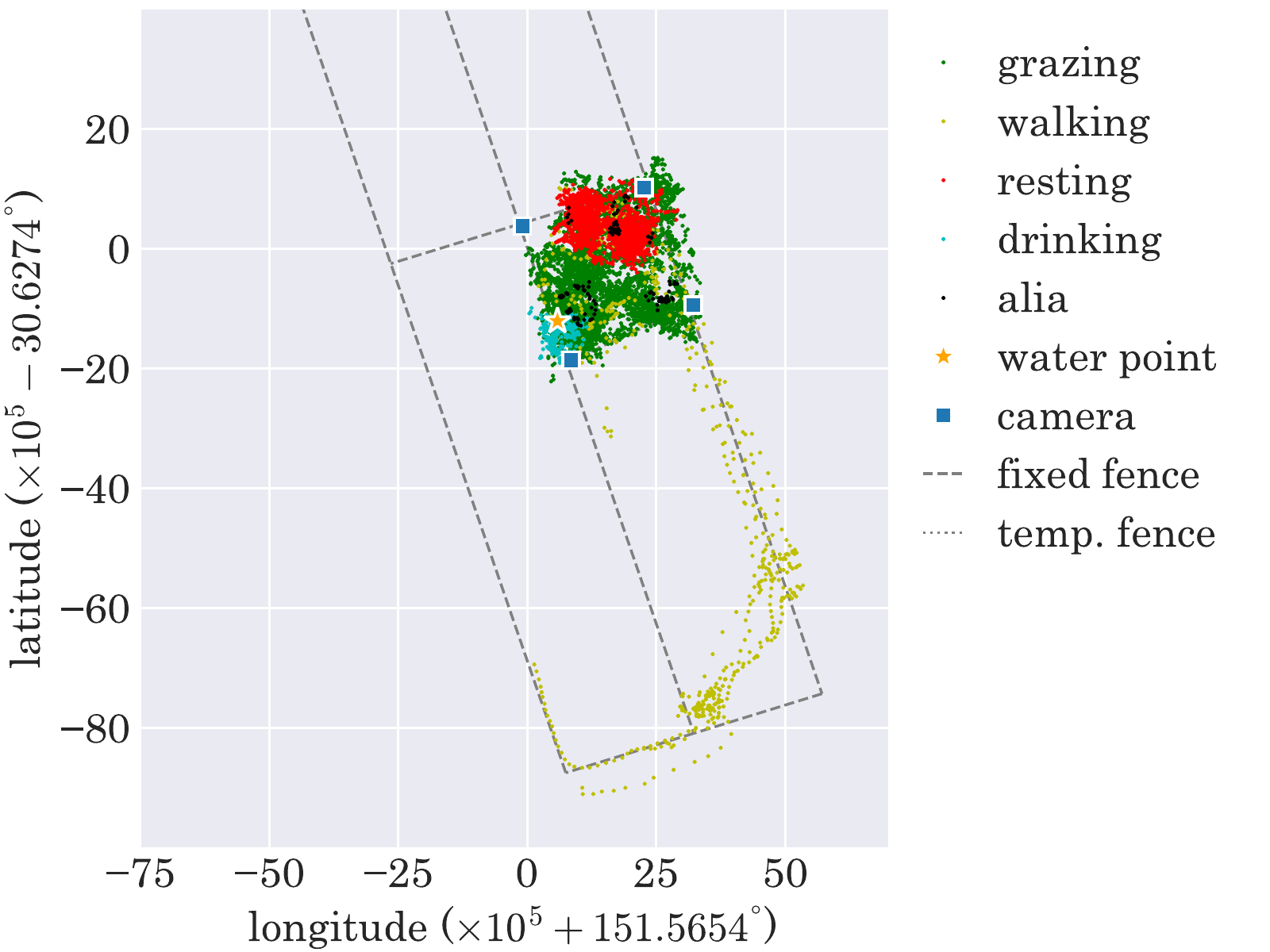}\label{26}}
    \subfigure[27/March/2020]{\includegraphics[scale=.545]{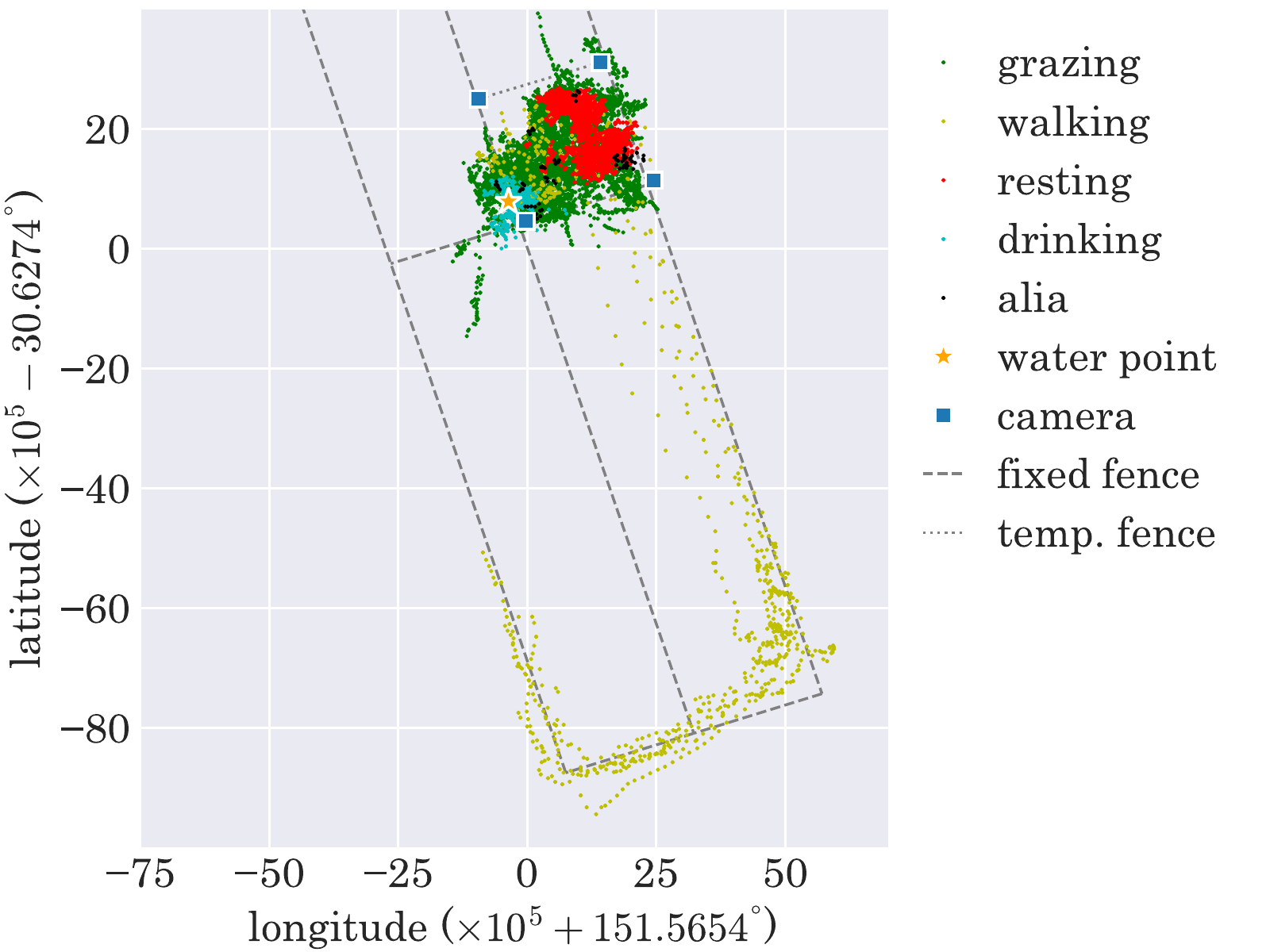}\label{27}}
    \end{multicols}
    \caption{\small{The GNSS-estimated position coordinates of the datapoints in the Arm20c dataset for every calendar day of the experiment leading to the dataset. Each position point is colored according to its associated behavior class.}}
   \label{gps_fixes_days}
\end{figure}

In Fig.~\ref{gps_fixes_days}, we plot the GNSS-estimated positions (longitude and latitude coordinates) of all datapoints for the Arm20c dataset. We plot the estimated positions in separate figures for each calendar day and color them according to their corresponding behavior class label. During every day of the data collection experiment, we changed the location of the water point and used different 25m-by-25m sections of the paddocks to maintain the cattle within the field of view of the cameras utilized to record the footage of the experiment. Therefore, in Fig.~\ref{gps_fixes_days}, we also indicate the location of the water point and those of the cameras installed at the corners of each daily experiment area as well as the permanent and temporary fence lines outlining the area used for the experiment each day.

\begin{figure}
    \centering
    \subfigure[grazing]{\includegraphics[scale=.545]{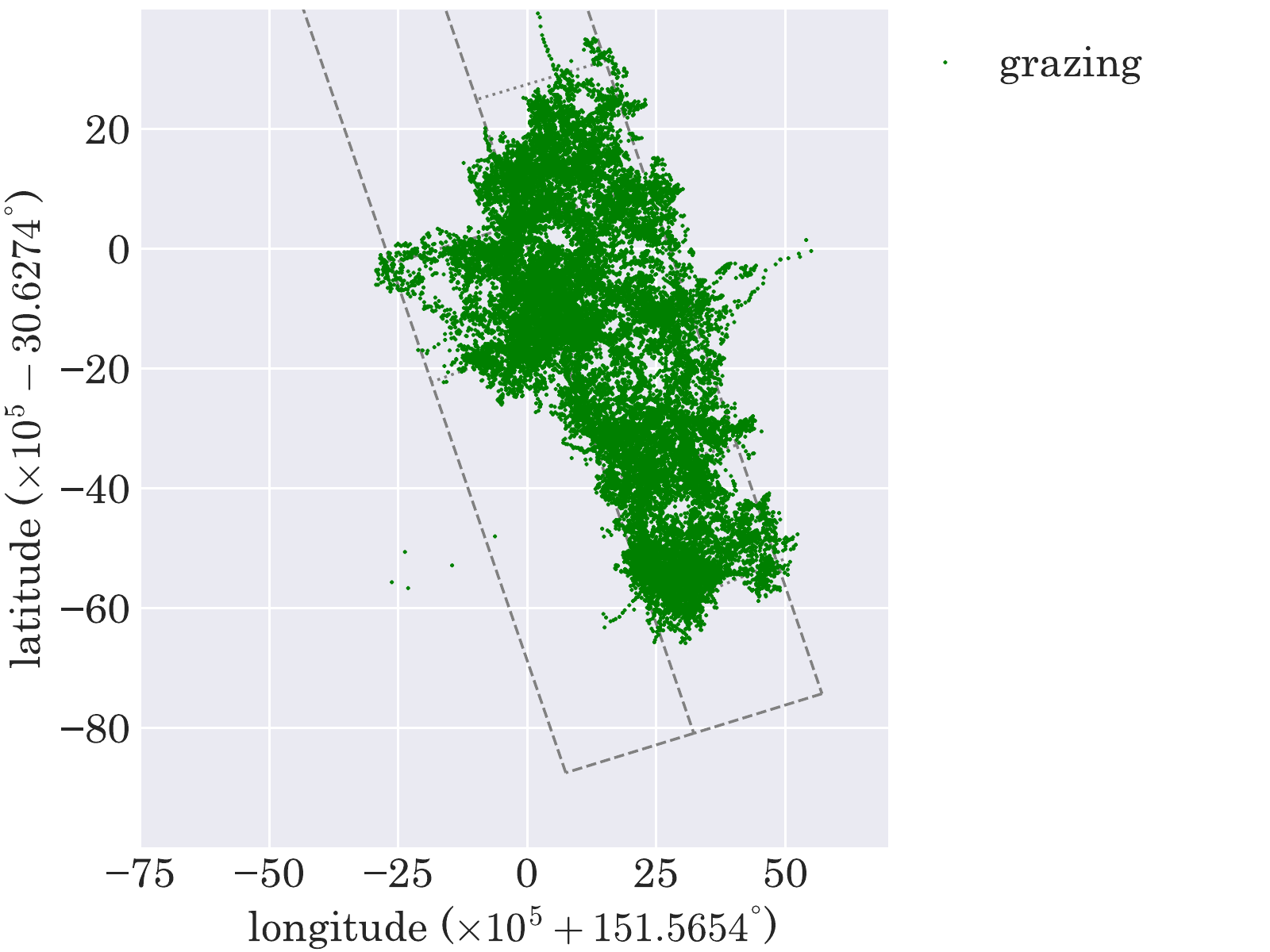}\label{fixes_grazing}}
    \subfigure[resting]{\includegraphics[scale=.545]{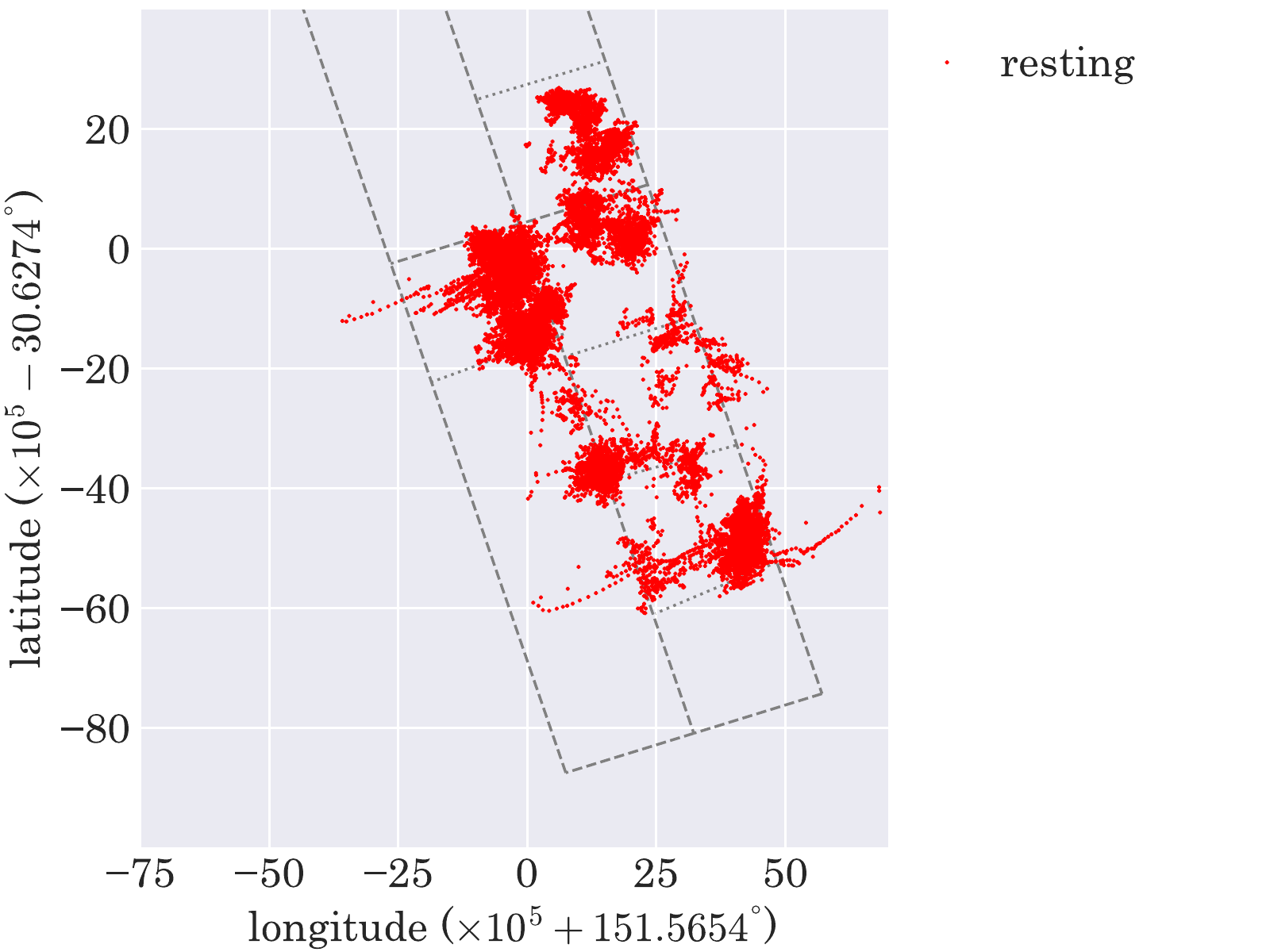}\label{fixes_resting}}
    \subfigure[walking, drinking, and alia]{\includegraphics[scale=.545]{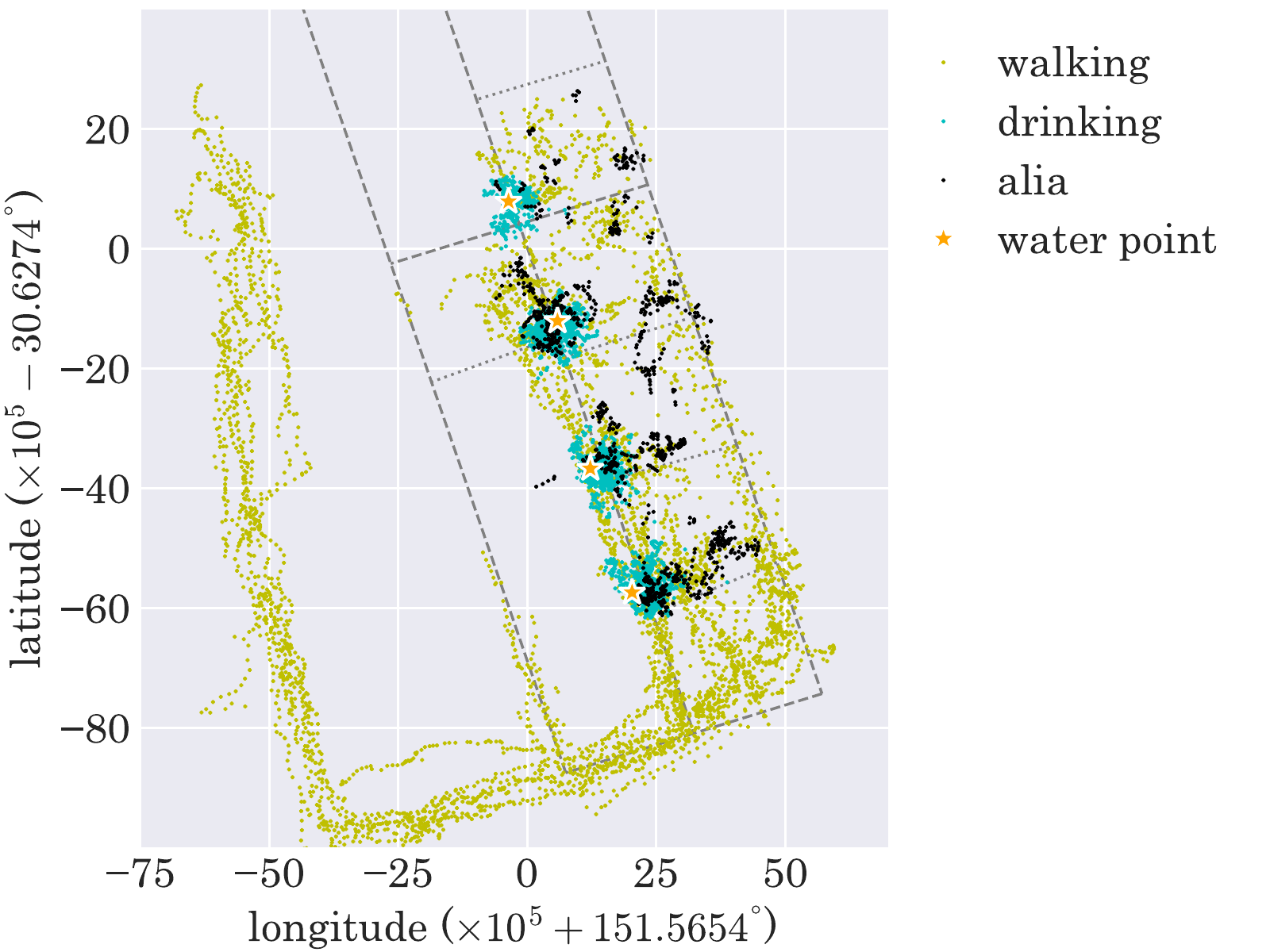}\label{fixed_rare}}
    \caption{\small{The GNSS-estimated position coordinates of the datapoints in the Arm20c dataset. Each position point is colored according to its associated behavior class.}}
   \label{gps_fixes}
\end{figure}

In Fig.~\ref{gps_fixes}, we plot the GNSS-estimated positions for all datapoints of the Arm20c dataset and the entire duration of the experiment. For clarity, we plot the estimated positions in three different figures corresponding to the grazing behavior, the resting behavior, and the remaining rare behaviors, i.e., walking, drinking, and alia.

As evident from Figs.~\ref{gps_fixes_days} and~\ref{gps_fixes} in agreement with intuition, the positions of the animals alone may not carry significant information about their behavior as they exhibit several behaviors such as grazing, resting, walking, and grooming at various locations on the paddock. However, the animals' distance from the places of interest such as the water point is more likely to contain discriminative information beneficial for behavior classification. It is clear from Fig.~\ref{fixed_rare} that the animals are in the proximity of the water point wherever their behavior is annotated as drinking. Note that the positions at which the animals' behavior is annotated as walking include the paths that connect the cattle pen to the daily experiment areas. The paths are distinctly visible in Figs.~\ref{gps_fixes_days} and~\ref{fixed_rare}.

\begin{figure}
    \centering
    \subfigure[DtWP]{\includegraphics[scale=.545]{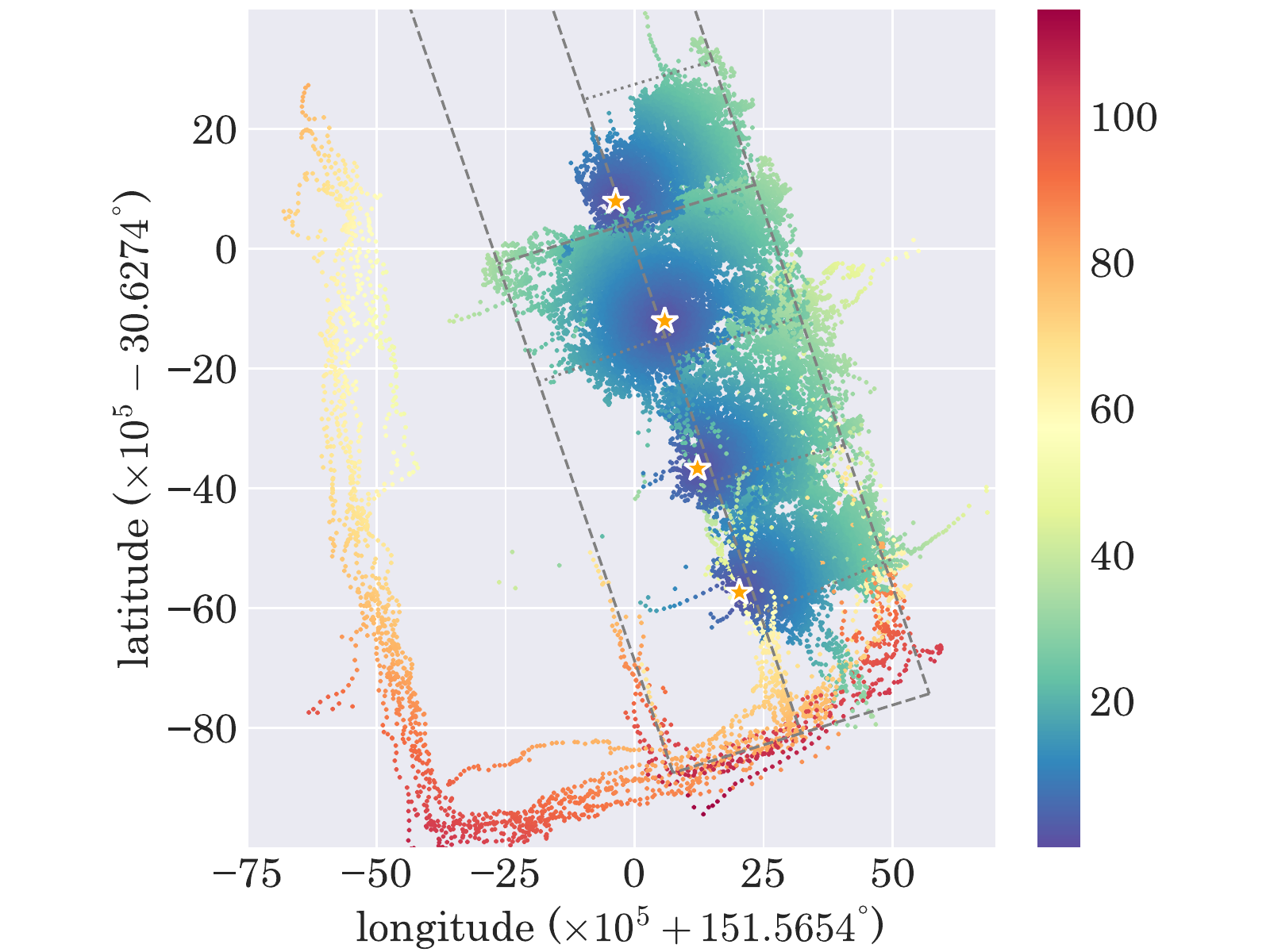}\label{gps_dtwp}}
    \subfigure[speed]{\includegraphics[scale=.545]{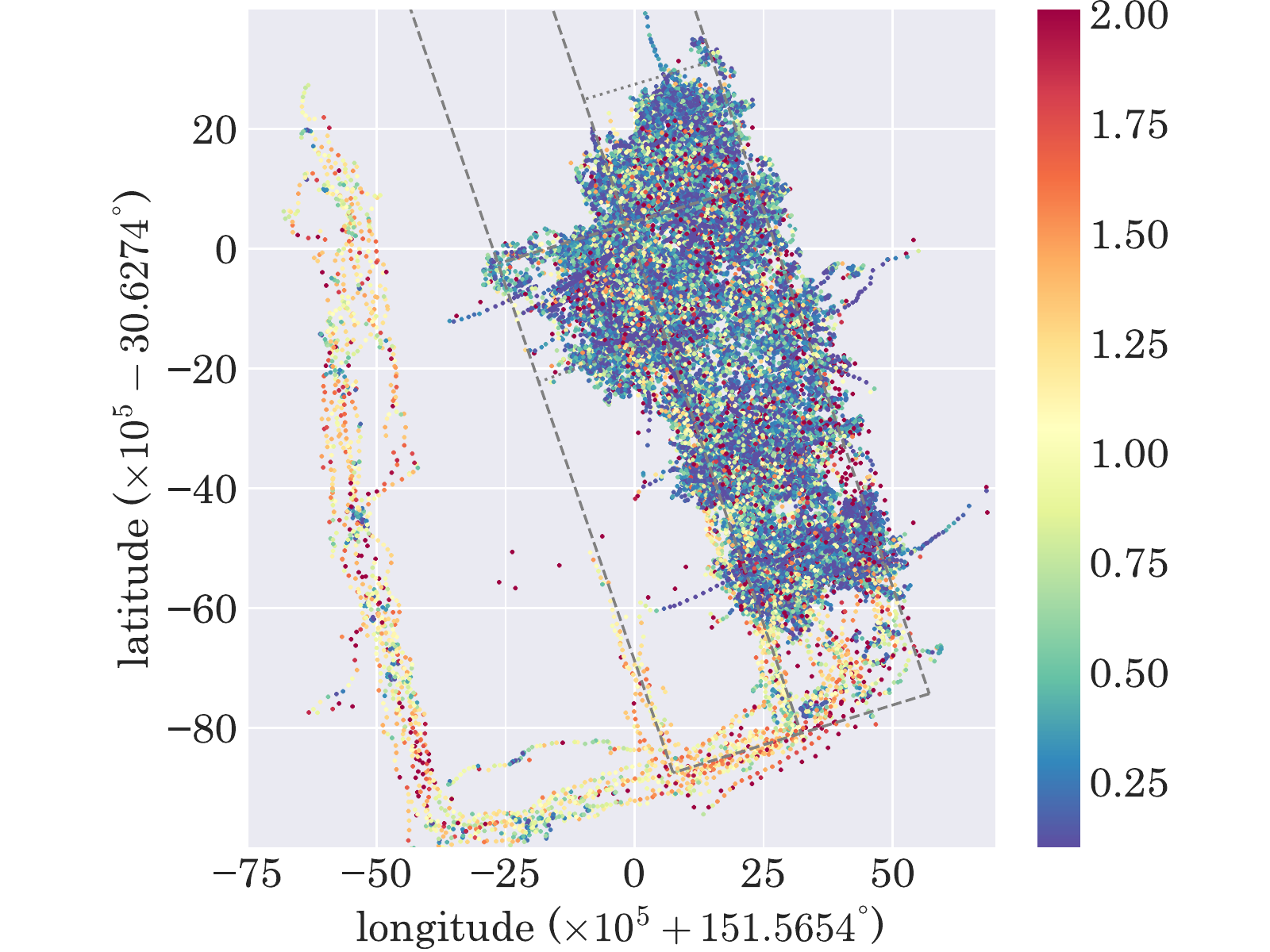}\label{gps_speed}}
    \subfigure[error]{\includegraphics[scale=.545]{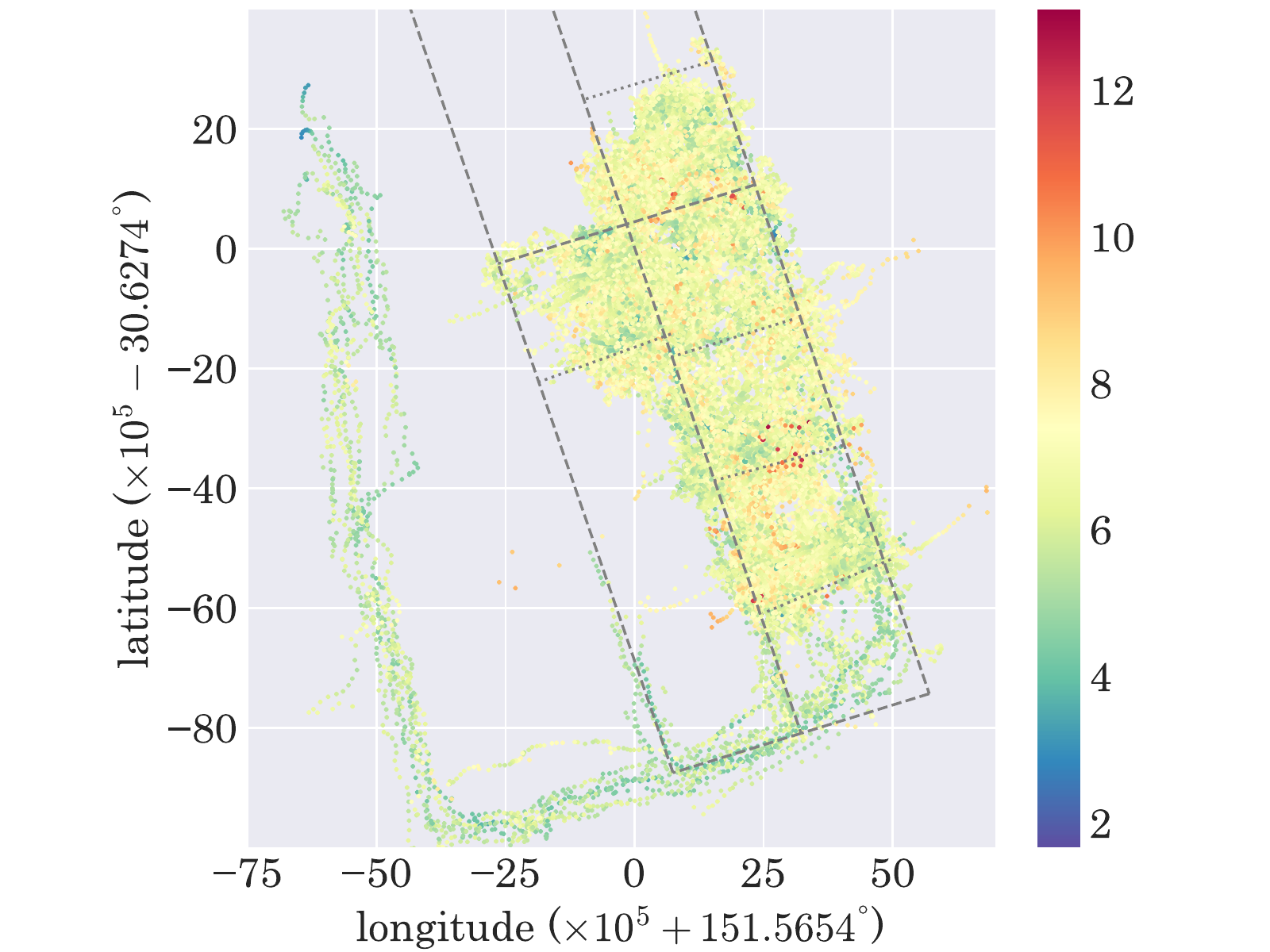}\label{gps_error}}
    \caption{\small{The GNSS-estimated position coordinates of the datapoints in the Arm20c dataset colored according to their corresponding values of DtWP, speed, and error.}}
   \label{gps_features}
\end{figure}

\begin{figure}
    \centering
    \subfigure[DtWP]{\includegraphics[scale=.545]{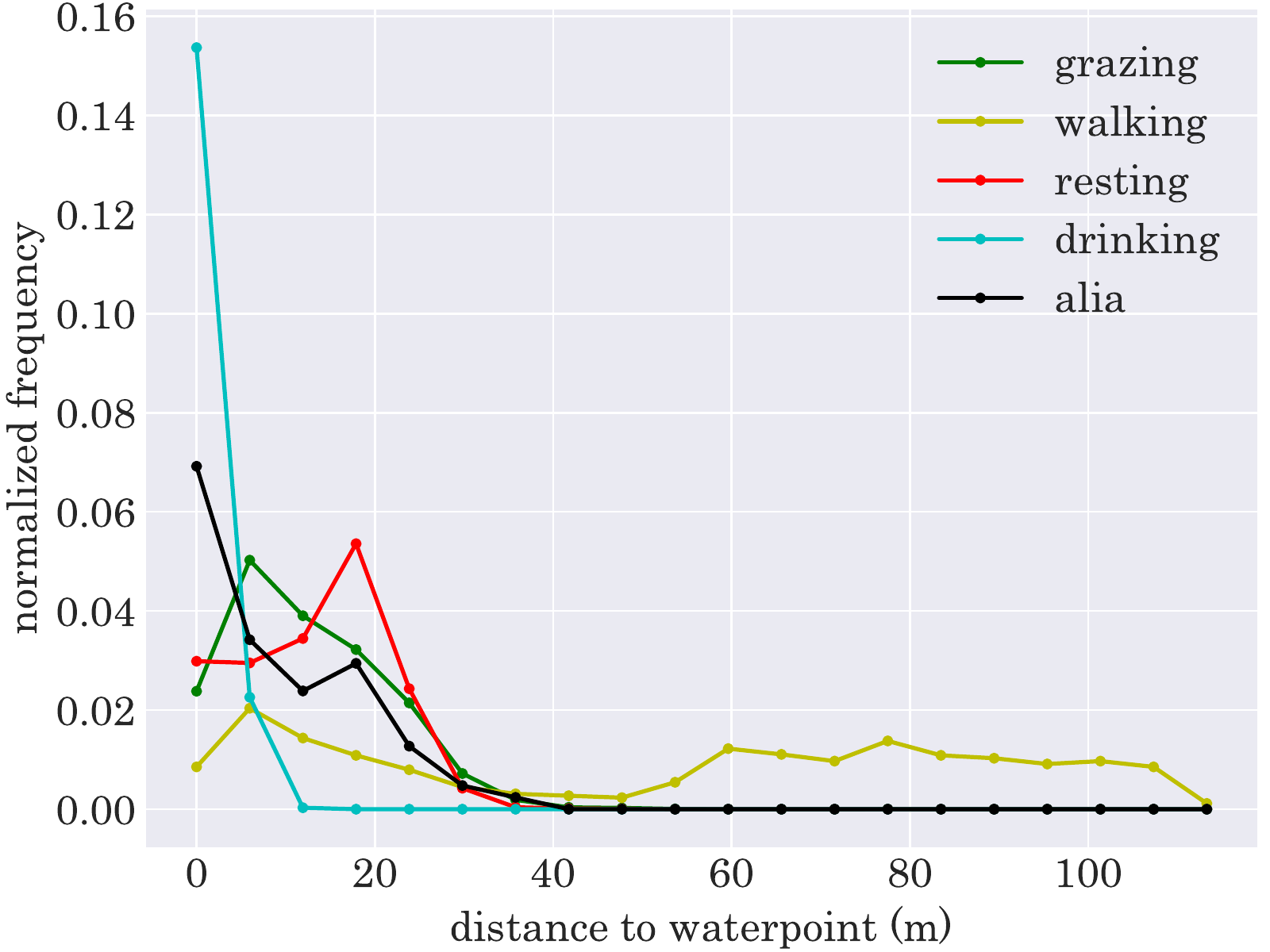}\label{hist_d2wp}}
    \subfigure[speed]{\includegraphics[scale=.545]{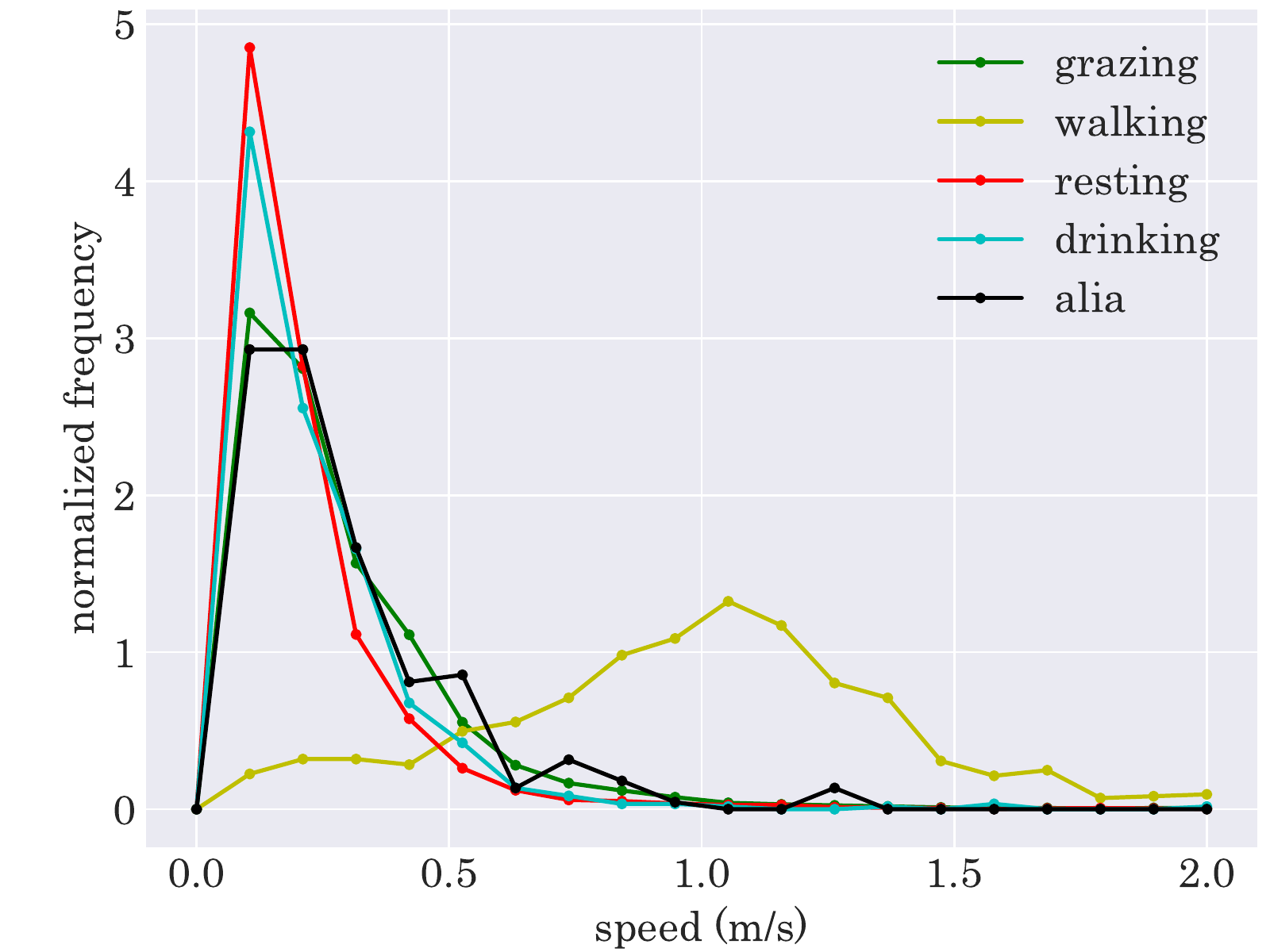}\label{hist_speed}}
    \subfigure[error]{\includegraphics[scale=.545]{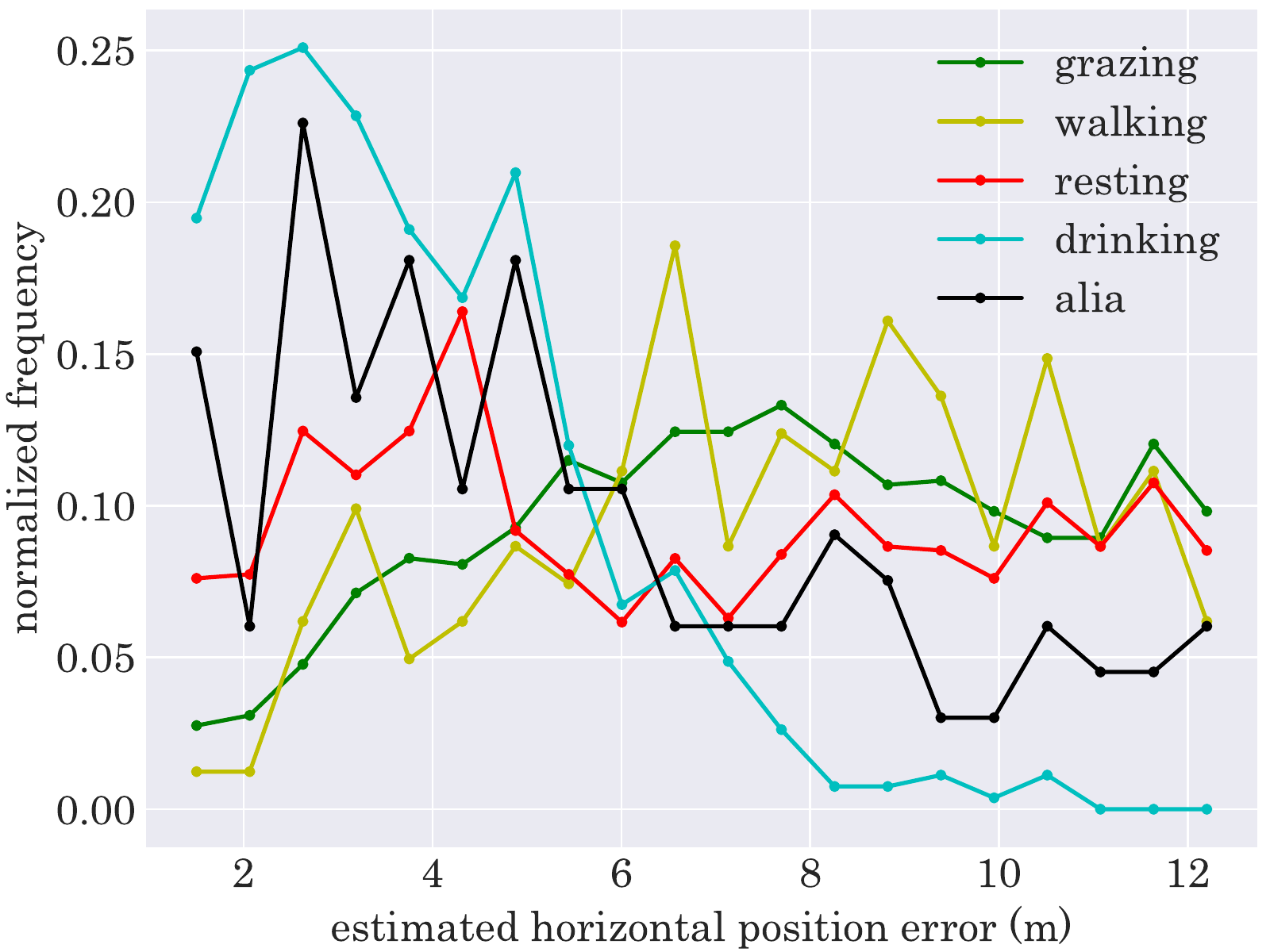}\label{hist_error}}
    \caption{\small{The histograms of the values of the GNSS features for the Arm20c dataset and each behavior class.}}
   \label{hists}
\end{figure}

To gain some insights into the values of the features calculated from the GNSS data and their correlation with the behavior class, in Fig.~\ref{gps_features}, we plot the GNSS-estimated positions and color them according to their corresponding values of the GNSS features, i.e., DtWP, speed, and error, for all datapoints of the Arm20c dataset. In addition, we plot the histograms of the values of the GNSS features for each behavior class in Fig.~\ref{hists}. The histograms are also related to the Arm20c dataset. However, those for the Arm20e dataset are qualitatively similar.

Figs.~\ref{gps_dtwp} and~\ref{hist_d2wp} show that the values of the DtWP feature are lower when the animals drink from the water point compared to when they exhibit other behaviors. It is also evident in Figs.~\ref{gps_speed} and~\ref{hist_speed} that the speed feature often has significantly higher values when the animals walk compared to when they perform other behaviors. Therefore, we can expect that the addition of these features to the features extracted from the accelerometry data should help improve the accuracy of behavior classification, especially regarding the uncommon behavior classes of walking and drinking. On the other hand, the values of the error feature do not appear to correlate with the behavior in any discernible manner, hence not carry any substantial information about the behavior directly. However, as we will show in section~\ref{abls}, the error feature can be informative for behavior classification, specifically, when used in conjunction with the other GNSS features.

\subsection{Classification performance}

We evaluate the classification accuracy of the considered algorithms and tune their hyperparameters via leave-one-animal-out cross-validation using our labeled datasets Arm20c and Arm20e. We calculate the cross-validated results by aggregating the results of all cross-validation folds where, in each fold, we use the datapoints associated with one animal for validation and the rest for training.

We use the Matthews correlation coefficient (MCC)~\citep{mcc} for evaluating the classification accuracy. MCC takes into account true and false positives and negatives and is meaningful even with highly imbalanced datasets. It has values between $-1$ and $+1$ where $+1$ is perfect prediction, $0$ is no better than random prediction, and $-1$ is perfect inverse prediction.

With the Arm20c dataset, we use six accelerometry features as described in section~\ref{acc_feat} while setting the values of the filter parameters to $\gamma_d=0.75$, $d\in\{x,y,z\}$, which we determine empirically through cross-validation. With the Arm20e dataset, we use nine accelerometry features, six of which are computed in the same way as in section~\ref{acc_feat} while setting $\gamma_d=0.75$, $d\in\{x,y,z\}$. The other three features are an extra set of $s_{id}$, $d\in\{x,y,z\}$, features with the associated filter parameter values of $\gamma_d=0.5$, $d\in\{x,y,z\}$. These extra features provide more information about the intensity of the animal’s body motion, particularly, at the higher parts of the associated frequency spectrum as with $\gamma_d=0.5$ lower frequencies are filtered out more aggressively compared to with $\gamma_d=0.75$. This can be noted in~\citep[Fig. 4]{RA21}. The supplementary information provided by these additional features helps improve the classification accuracy with the Arm20e dataset.

We implement the MLP classifiers using the scikit-learn Python package\footnote{\url{https://scikit-learn.org/}}. We utilize the L-BFGS algorithm~\citep{lbfgs} to optimize the cross-entropy loss associated with the classification. We set the number of units in the hidden layer of each MLP to the average of the numbers of its input and output neurons, which are the corresponding features and classes, rounded upward. We tune the $\ell_2$-norm regularization hyperparameter of each MLP for each dataset in conjunction with our leave-one-animal-out cross-validation process. To train each MLP, we run the L-BFGS optimization algorithm for maximum $10,000$ iterations.

We initialize the model parameters (MLP weights and biases) of each classifier randomly. As the classification accuracy may vary depending on the initial values of the parameters, we repeat each cross-validation evaluation $100$ times using independently selected random initial parameter values and compute the average and standard deviation of the MCC values over the trials. Therefore, when reporting MCC values, we present the mean over $100$ independent runs followed by the associated standard-deviation separated by the $\pm$ sign.

In Tables~\ref{mcc-c} and~\ref{mcc-e}, we present the cross-validated classification accuracy results, in terms of MCC, for the Arm20c and Arm20e datasets, respectively. The tables include the MCC values for each behavior class as well as the overall MCC values. The results are for the classifier that only uses the GNSS features (GNSS), the classifier that only uses the accelerometry features (Acc), the classifier that uses the concatenated GNSS and accelerometry features (FC), and the classifier that fuses the posterior class probabilities predicted by the GNSS and Acc classifiers (PF).

\begin{table} \footnotesize
\caption{The leave-one-animal-out cross-validated MCC values of the considered algorithms for the Arm20c dataset and each behavior class. The MCC values are averaged over $100$ independent runs with random initialization.\\}
\label{mcc-c}
\centering
\begin{tabular}{|l|cccc|}
\hline
 \backslashbox{behavior}{algorithm} & GNSS & Acc & FC & PF\\
\hline
grazing  & $0.4348\pm0.0070$ & $0.9011\pm0.0031$ & $0.9252\pm0.0040$ & $0.9227\pm0.0037$\\
walking  & $0.7708\pm0.0130$ & $0.8361\pm0.0048$ & $0.8940\pm0.0101$ & $0.8900\pm0.0131$\\
resting  & $0.3931\pm0.0081$ & $0.9000\pm0.0063$ & $0.8956\pm0.0059$ & $0.9050\pm0.0046$\\
drinking & $0.1952\pm0.0148$ & $0.6444\pm0.0186$ & $0.7508\pm0.0172$ & $0.7679\pm0.0114$\\
alia     & $-$               & $0.2785\pm0.0337$ & $0.3755\pm0.0292$ & $0.3256\pm0.0319$\\
\hline
overall  & $0.4375\pm0.0063$ & $0.8594\pm0.0045$ & $0.8829\pm0.0048$ & $0.8847\pm0.0043$\\
\hline
\end{tabular}
\end{table}

\begin{table} \footnotesize
\caption{The leave-one-animal-out cross-validated MCC values of the considered algorithms for the Arm20e dataset and each behavior class. The MCC values are averaged over $100$ independent runs with random initialization.\\}
\label{mcc-e}
\centering
\begin{tabular}{|l|cccc|}
\hline
\backslashbox{behavior}{algorithm} & GNSS & Acc & FC & PF\\
\hline
grazing    & $0.3516\pm0.0063$ & $0.7659\pm0.0070$ & $0.8122\pm0.0060$ & $0.8140\pm0.0055$\\
walking    & $0.7545\pm0.0118$ & $0.6607\pm0.0155$ & $0.8204\pm0.0135$ & $0.8305\pm0.0098$\\
resting    & $0.3138\pm0.0084$ & $0.7229\pm0.0069$ & $0.7431\pm0.0074$ & $0.7494\pm0.0065$\\
drinking   & $0.1179\pm0.0201$ & $0.0841\pm0.0287$ & $0.4352\pm0.0185$ & $0.4487\pm0.0227$\\
alia       & $-$ & $0.1191\pm0.0444$ & $0.1717\pm0.0370$ & $0.1192\pm0.0296$\\
\hline
overall    & $0.3713\pm0.0057$ & $0.6916\pm0.0064$ & $0.7475\pm0.0059$ & $0.7532\pm0.0051$\\
\hline
\end{tabular}
\end{table}

The results in Tables~\ref{mcc-c} and \ref{mcc-e} show that combining the GNSS and accelerometry features improves the classification accuracy appreciably while fusing the posterior probabilities produced by the classifiers that use GNSS and accelerometry features separately appears to outperform the classifier that uses the concatenated features. We also observe that the performance enhancement offered by the GNSS features is more prominent for the uncommon behaviors of walking and drinking, especially with the Arm20e dataset. This is not surprising since, as discussed in section~\ref{gnss_anal}, the DtWP and speed features are expected to provide additional information that can help better distinguish the drinking and walking instances. Note that using the GNSS features alone does not lead to any meaningful classification of the drinking behavior. However, when these features are used in conjunction with the accelerometry features, the accuracy of classifying the drinking behavior improves substantially.

\subsection{Ablation study}\label{abls}

In Tables~\ref{abls-mcc-c} and~\ref{abls-mcc-e}, we provide the cross-validated MCC values of the PF algorithm when different combinations of the GNSS features are used alongside the accelerometry features for both considered datasets and each behavior class. It is clear from the results that, among the considered GNSS features, DtWP is the most informative one regarding the classification of all considered behaviors. The other two GNSS features, speed and error, are beneficial to behavior classification with Arm20e dataset. However, they do not seem to offer any noticeable advantage with the Arm20c dataset. This may be because of higher complexity of classifying animal behavior using the ear tag accelerometry data that makes it more amenable to exploit any extra knowledge imparted by additional features such as speed and error. Moreover, the error feature appears to be more helpful when it is used together with the DtWP feature. This is possibly due to the error feature carrying some information about how reliable the DtWP feature value of each datapoint is.

The results of Tables~\ref{abls-mcc-c} and~\ref{abls-mcc-e} show that the DtWP feature helps improve the accuracy of classifying the drinking behavior significantly with both considered datasets. It also helps with the classification of the walking behavior in both datasets. In addition, using the speed feature enhances the classification accuracy of the walking behavior with the Arm20e dataset. All considered GNSS features can benefit the classification of the grazing and resting behaviors as well, although to a lesser extent compared to the drinking and walking behaviors and to varying degrees given different datasets.

\begin{landscape}

\begin{table} \footnotesize
\caption{The leave-one-animal-out cross-validated MCC values of the PF algorithm when using different combinations of the GNSS features in the GNSS-based classifier for the Arm20c dataset and each behavior class.\\}
\label{abls-mcc-c}
\centering
\begin{tabular}{|l|c|ccc|ccc|c|}
\hline
\backslashbox{behavior}{GNSS features} & all & \begin{tabular}{@{}c@{}}DtWP\\error\end{tabular} & \begin{tabular}{@{}c@{}}DtWP\\speed\end{tabular} & \begin{tabular}{@{}c@{}}speed\\error\end{tabular} & DtWP & speed & error & none\\
\hline
grazing  & $0.9227\pm0.0037$ & $0.9222\pm0.0031$ & $0.9197\pm0.0049$ & $0.9038\pm0.0033$ & $0.9216\pm0.0040$ & $0.8938\pm0.0036$ & $0.9033\pm0.0038$ & $0.9011\pm0.0031$\\
walking  & $0.8900\pm0.0131$ & $0.8898\pm0.0044$ & $0.8891\pm0.0145$ & $0.8453\pm0.0100$ & $0.8889\pm0.0035$ & $0.8391\pm0.0086$ & $0.8196\pm0.0061$ & $0.8361\pm0.0048$\\
resting  & $0.9050\pm0.0046$ & $0.9041\pm0.0050$ & $0.9022\pm0.0064$ & $0.8950\pm0.0076$ & $0.9040\pm0.0050$ & $0.8982\pm0.0072$ & $0.8968\pm0.0074$ & $0.9000\pm0.0063$\\
drinking & $0.7679\pm0.0114$ & $0.7664\pm0.0129$ & $0.7623\pm0.0156$ & $0.6442\pm0.0243$ & $0.7674\pm0.0125$ & $0.6453\pm0.0190$ & $0.6460\pm0.0230$ & $0.6444\pm0.0186$\\
alia     & $0.3256\pm0.0319$ & $0.3096\pm0.0331$ & $0.3045\pm0.0272$ & $0.3142\pm0.0292$ & $0.2895\pm0.0267$ & $0.2980\pm0.0291$ & $0.2696\pm0.0346$ & $0.2785\pm0.0337$\\
\hline
overall  & $0.8847\pm0.0043$ & $0.8845\pm0.0037$ & $0.8841\pm0.0053$ & $0.8593\pm0.0058$ & $0.8819\pm0.0038$ & $0.8597\pm0.0055$ & $0.8587\pm0.0056$ & $0.8594\pm0.0045$\\
\hline
\end{tabular}
\end{table}

\begin{table} \footnotesize
\caption{The leave-one-animal-out cross-validated MCC values of the PF algorithm when using different combinations of the GNSS features in the GNSS-based classifier for the Arm20e dataset and each behavior class.\\}
\label{abls-mcc-e}
\centering
\begin{tabular}{|l|c|ccc|ccc|c|}
\hline
\backslashbox{behavior}{GNSS features} & all & \begin{tabular}{@{}c@{}}DtWP\\error\end{tabular} & \begin{tabular}{@{}c@{}}DtWP\\speed\end{tabular} & \begin{tabular}{@{}c@{}}speed\\error\end{tabular} & DtWP & speed & error & none\\
\hline
grazing  & $0.8140\pm0.0055$ & $0.8104\pm0.0063$ & $0.7982\pm0.0059$ & $0.7956\pm0.0055$ & $0.7984\pm0.0077$ & $0.7739\pm0.0071$ & $0.7884\pm0.0060$ & $0.7659\pm0.0070$\\
walking  & $0.8305\pm0.0098$ & $0.8062\pm0.0084$ & $0.7957\pm0.0094$ & $0.7702\pm0.0091$ & $0.7947\pm0.0097$ & $0.7195\pm0.0080$ & $0.6885\pm0.0112$ & $0.6607\pm0.0155$\\
resting  & $0.7494\pm0.0065$ & $0.7373\pm0.0074$ & $0.7449\pm0.0069$ & $0.7373\pm0.0095$ & $0.7333\pm0.0072$ & $0.7345\pm0.0099$ & $0.7243\pm0.0085$ & $0.7229\pm0.0069$\\
drinking & $0.4487\pm0.0227$ & $0.4444\pm0.0268$ & $0.4572\pm0.0226$ & $0.1149\pm0.0350$ & $0.4546\pm0.0239$ & $0.0938\pm0.0318$ & $0.0940\pm0.0333$ & $0.0841\pm0.0287$\\
alia     & $0.1192\pm0.0296$ & $0.1072\pm0.0372$ & $0.1029\pm0.0266$ & $0.1152\pm0.0379$ & $0.0990\pm0.0359$ & $0.1166\pm0.0368$ & $0.1189\pm0.0423$ & $0.1191\pm0.0444$\\
\hline
overall  & $0.7532\pm0.0051$ & $0.7443\pm0.0058$ & $0.7411\pm0.0053$ & $0.7228\pm0.0069$ & $0.7371\pm0.0063$ & $0.7058\pm0.0072$ & $0.7042\pm0.0061$ & $0.6916\pm0.0064$\\
\hline
\end{tabular}
\end{table}

\end{landscape}

\subsection{Computational complexity}

Table~\ref{comp} shows the number of parameters of the considered animal behavior classification algorithms as well as the number of different arithmetic operations required by their associated MLP classifier(s). The numbers include the additional parameters and operations required by the PF algorithm for fusing the posterior probabilities but exclude the parameters and operations required for feature extraction as they are the same for the FC and PF algorithms. As we see in Table~\ref{comp}, the PF algorithm requires fewer parameters and arithmetic operations compared with the FC algorithm. The savings offered by the PF algorithm over the FC algorithm are higher with the Arm20e dataset as we extract more accelerometry features from this dataset.

\begin{table} \footnotesize
\caption{The number of parameters and the number of arithmetic operations required by the classification part of the considered animal behavior classification algorithms for both considered datasets.\\}
\label{comp}
\centering
\begin{tabular}{|l|c|ccc|ccc|}
\hline
\vspace{-14pt}
& & \multicolumn{3}{c|}{Arm20c} &\multicolumn{3}{c|}{Arm20e}\\
\backslashbox{complexity}{dataset\\algorithm} & GNSS & Acc & FC & PF & Acc & FC & PF\\
\hline
parameters              & $32$ & $55$  & $98$  & $92$  & $98$  & $153$ & $135$\\
\hline
multiplications         & $32$ & $55$  & $98$  & $87$  & $98$  & $153$ & $130$\\
additions/subtractions  & $32$ & $55$  & $98$  & $97$  & $98$  & $153$ & $140$\\
ReLU operations         & $4$  & $5$   & $7$   & $9$   & $7$   & $9$   & $11$\\
\hline
sum of operations        & $68$ & $115$ & $203$ & $193$ & $203$ & $315$ & $281$\\
\hline
\end{tabular}
\end{table}

\subsection{Feature-space visualization}

\begin{figure}
    \centering
    \subfigure[Accelerometry features.]{\includegraphics[scale=.545]{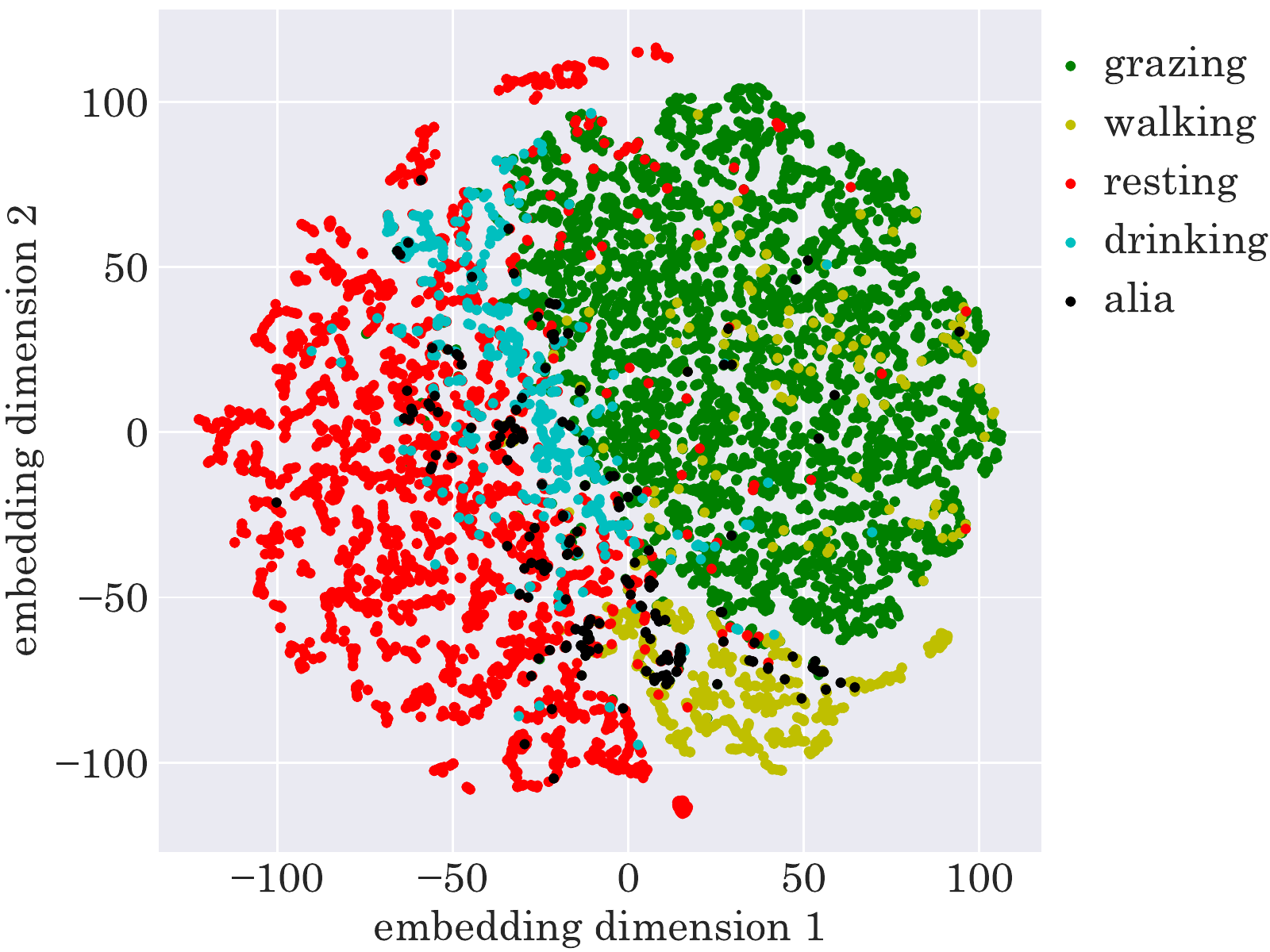}\label{tsne-c-acc}}
    \subfigure[GNSS features.]{\includegraphics[scale=.545]{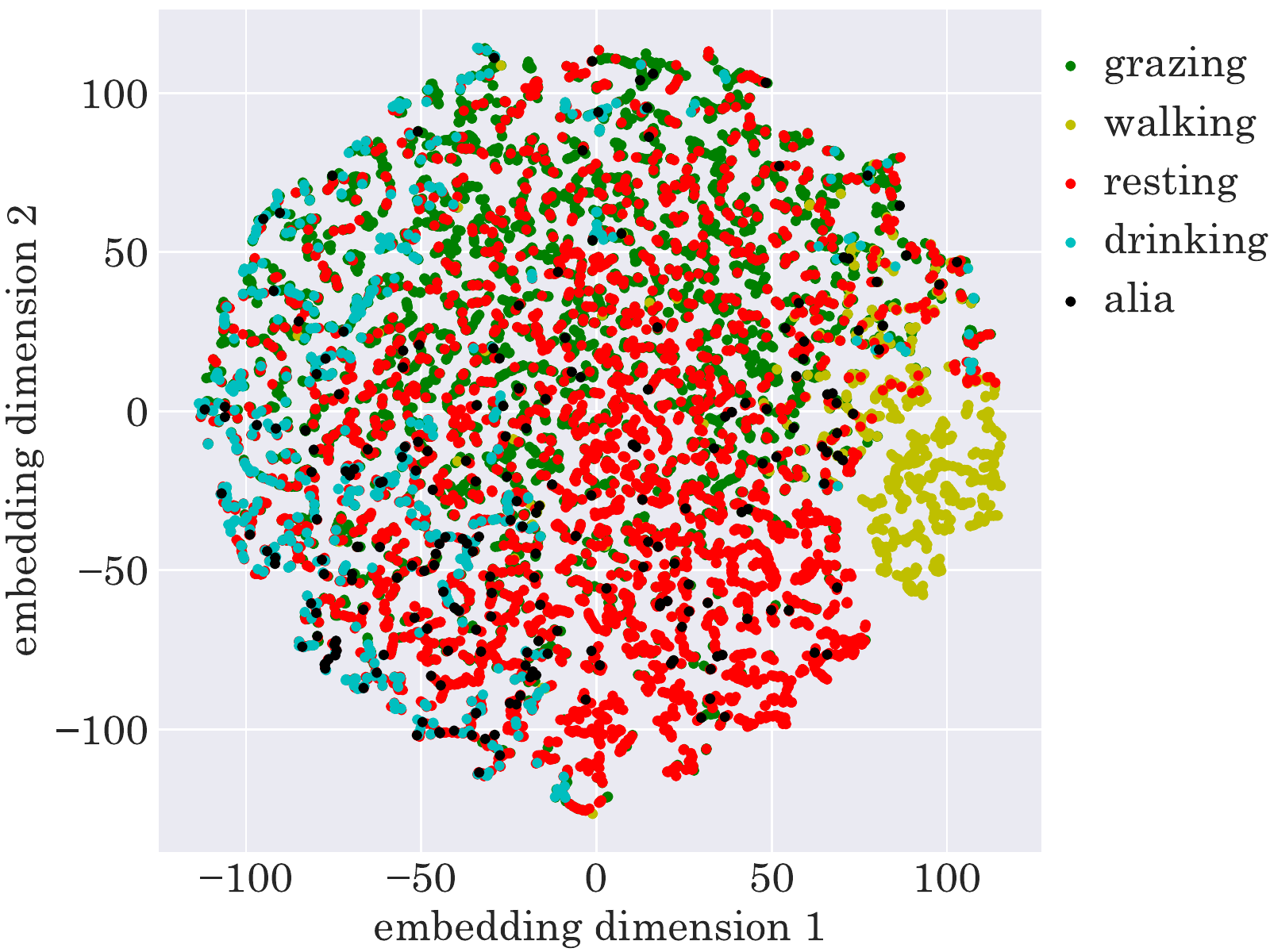}\label{tsne-c-gnss}}
    \subfigure[Accelerometry and GNSS features.]{\includegraphics[scale=.545]{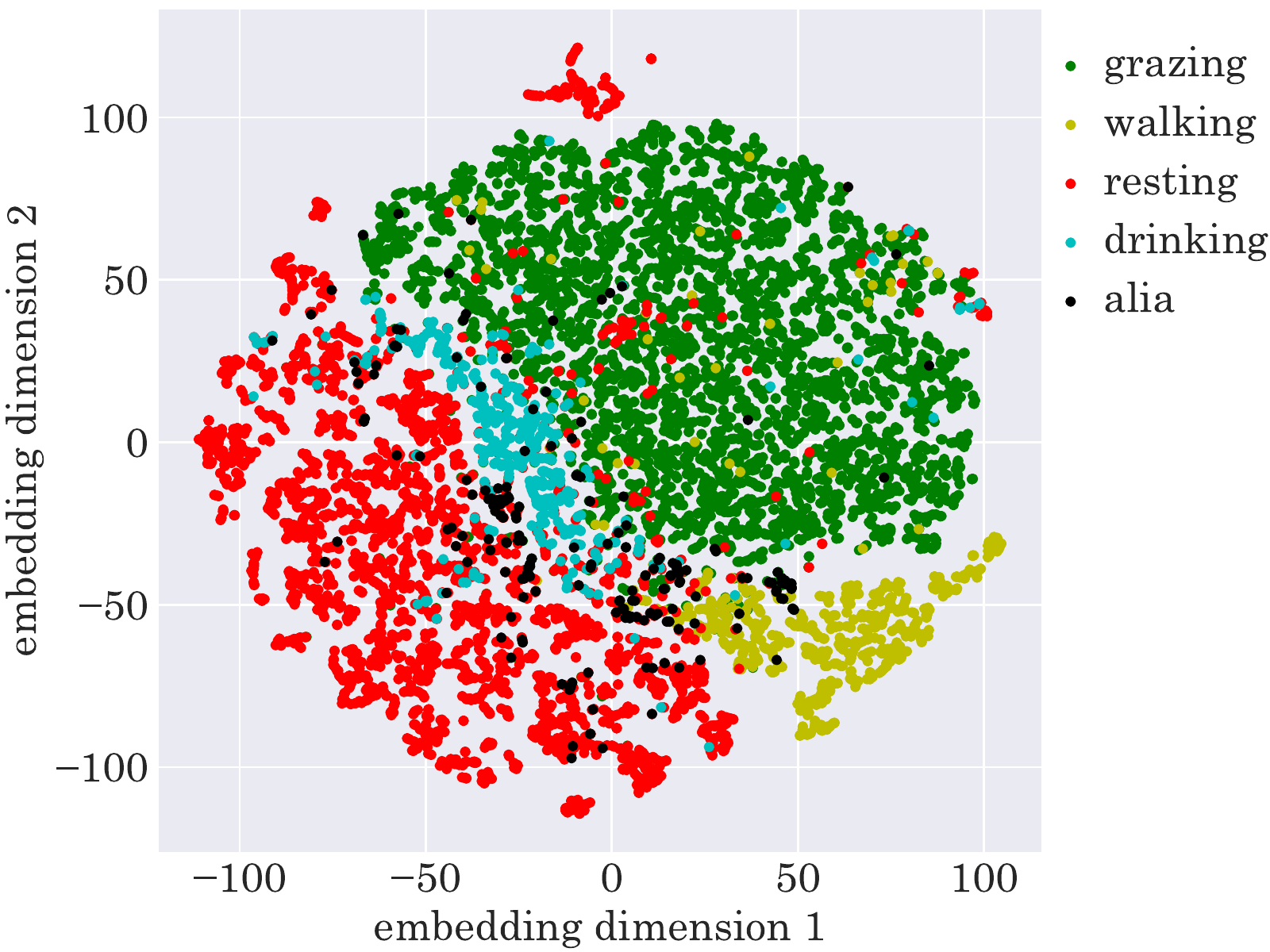}\label{tsne-c-both}}
    \caption{\small{The tSNE visualization of the feature subspace of the Arm20c dataset using the accelerometry features, GNSS features, or both.}}
    \label{tsne-c}
\end{figure}

\begin{figure}
    \centering
    \subfigure[Accelerometry features.]{\includegraphics[scale=.545]{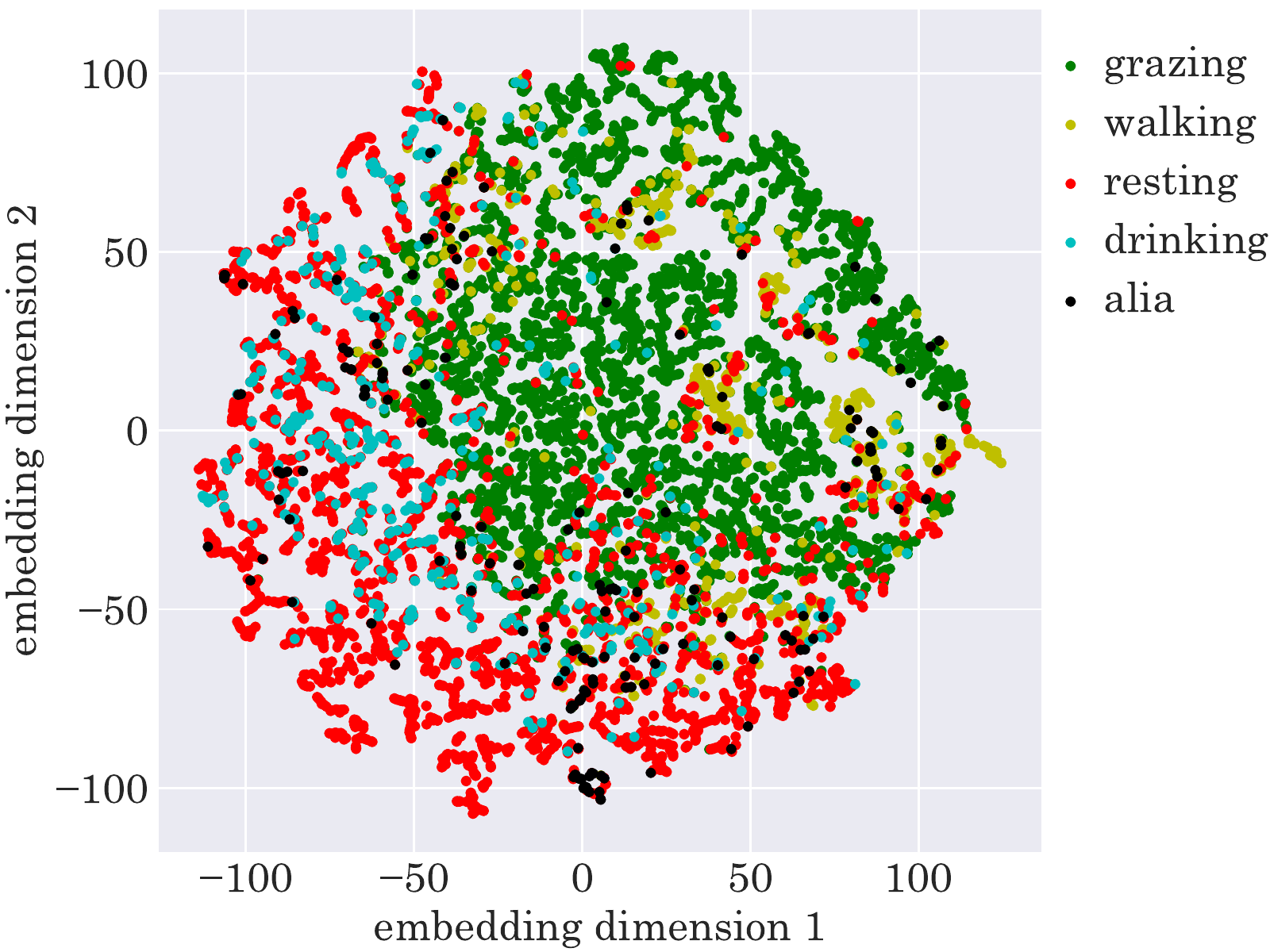}\label{tsne-e-acc}}
    \subfigure[GNSS features.]{\includegraphics[scale=.545]{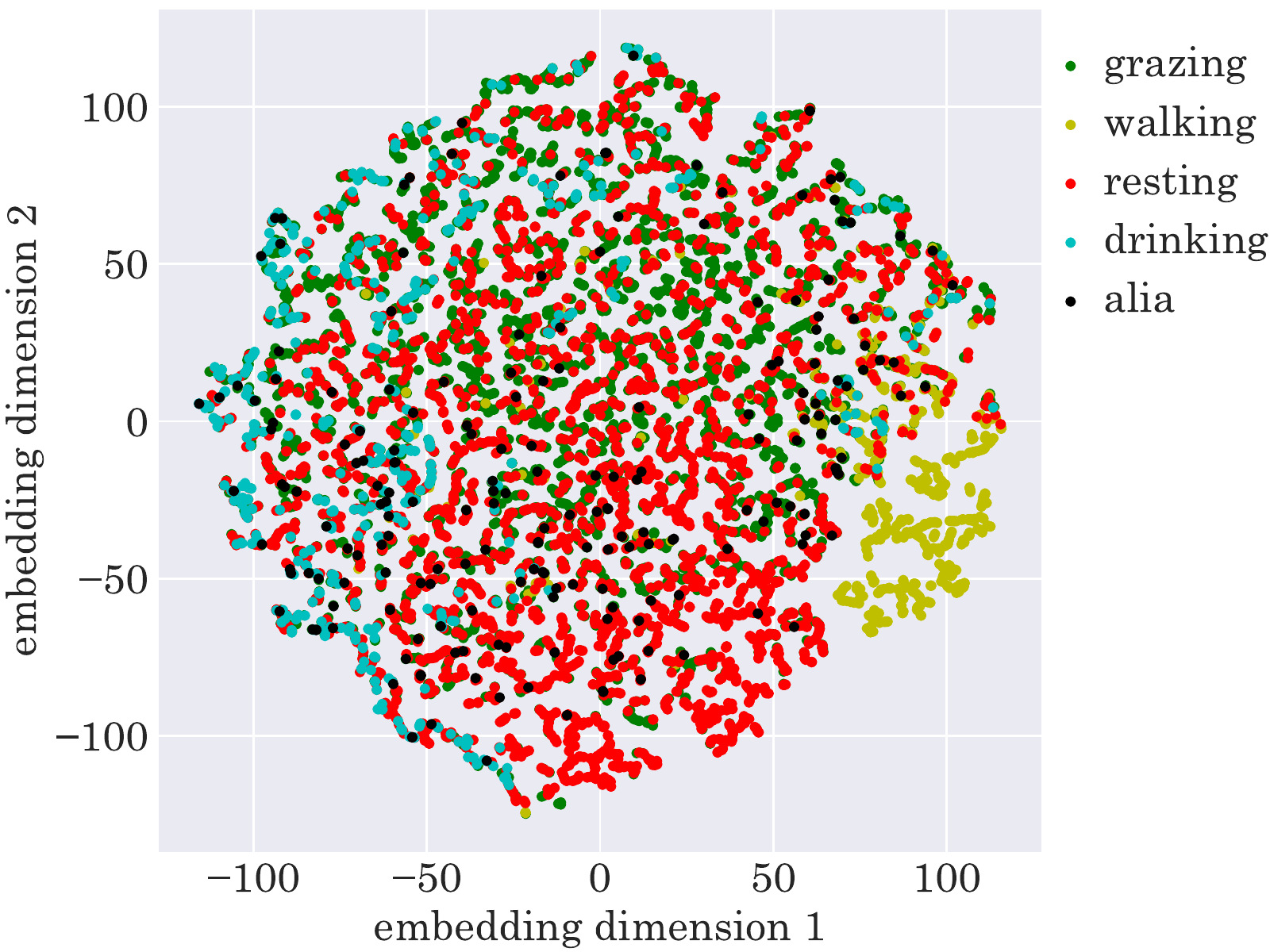}\label{tsne-e-gnss}}
    \subfigure[Accelerometry and GNSS features.]{\includegraphics[scale=.545]{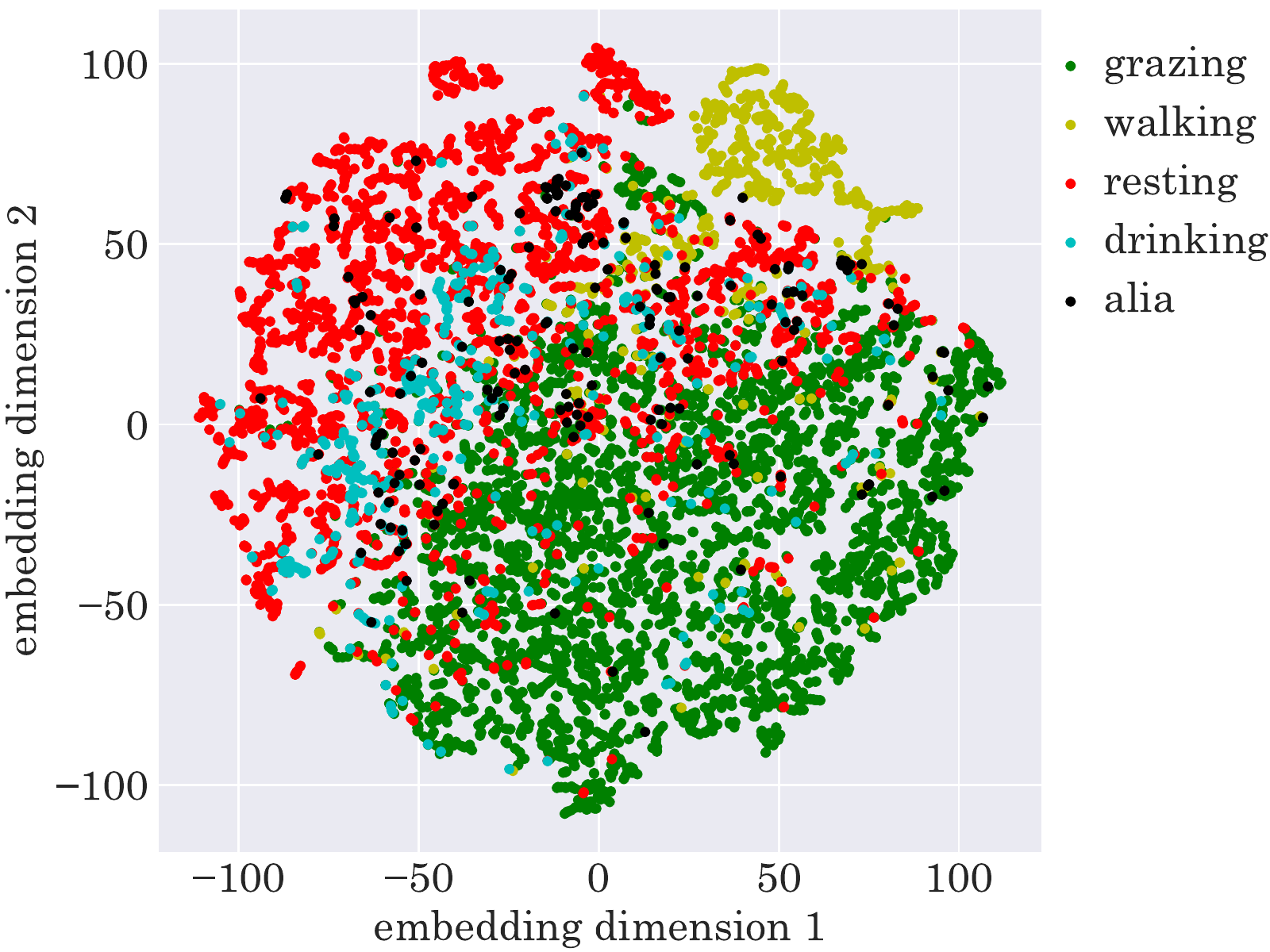}\label{tsne-e-both}}
    \caption{\small{The tSNE visualization of the feature subspace of the Arm20e dataset using the accelerometry features, GNSS features, or both.}}
    \label{tsne-e}
\end{figure}

To provide some insights into the benefits of combining the information available from the GNSS and accelerometry data, we visualize the feature subspace related to the Arm20c and Arm20e datasets using the $t$-distributed stochastic neighbor embedding (tSNE) algorithm~\citep{tSNE} in Figs.~\ref{tsne-c} and~\ref{tsne-e}, respectively. The figures include the visualizations of the feature subspace of each dataset when using the accelerometry features, GNSS features, or both. Each point represents a datapoint and is colored according to its corresponding behavior class. We calculate the features for the entire datasets using the parameters of the respective models trained for behavior classification. The tSNE algorithm preserves the local structure of the feature subspace while projecting it onto a lower-dimensional space with no guarantee on preserving the global structure of the data.

It is clear from Figs.~\ref{tsne-c} and~\ref{tsne-e} that augmenting the accelerometry features with the GNSS features enhances the clustering of datapoints belonging to the same class thus helps improve the classification accuracy. This is more noticeable for the less common behavior classes, i.e., walking and drinking.

\section{Discussion}

In this work, we consider two sensing modes of accelerometry and GNSS. However, our arguments and findings can be extended to any number or type of sensing modes. In addition, we use MLP classifiers while any other classification algorithm can also be used in principle. The only caveat is that, for the PF algorithm to be effective, the classifiers associated to individual sensing modes should either directly output class probabilities or output quantities that can be converted to posterior class probabilities. When using classifiers that do not intrinsically predict any probability distribution over classes, such as the support vector machine, one can use a calibration algorithm, such as Platt scaling~\citep{platt}, to transform the classification model output to posterior probabilities.

The assumption of the accelerometry and GNSS features being conditionally independent given the class label is akin to the underpinning assumption of the naive Bayes algorithm, albeit when the features are treated in blocks. However, the main difference between the PF algorithm and the naive Bayes algorithm is that the former combines the posterior probabilities while the latter combines the likelihoods.

Although the independence assumption may not be ideal, the PF algorithm has several advantages over the FC algorithm that make it more useful in practice. Particularly, statistical decoupling of the accelerometry and GNSS features means that each classifier can operate independently in a lower-dimensional parameter space. As a result, multimodal classification can be delegated to smaller single-modal classifiers reducing overall computational and memory requirements. This can also help mitigate problems that may arise with high-dimensional data such as the need for datasets that scale exponentially with the number of features.

Another important advantage of the PF algorithm over the FC algorithm is its ability to effectively handle instances when either of the accelerometry or GNSS data is corrupt or missing due to a fault or difference in the availability frequency. While the FC algorithm fails in such instances, the PF algorithm can still function and classify the behavior, although with reduced accuracy. This can be realized by detecting the missing/corrupt sensor data and outputting the predictions made by the classifier associated with the valid sensor data instead of preforming the posterior probability fusion. In the case of having more than two sensing modes, the contribution of the invalid sensor data can be ignored in the probability fusion stage. On the other hand, the FC algorithm may fail to make any meaningful prediction when any feature is missing or corrupt as it requires all feature values to estimate the posterior probabilities.

Fusing posterior probabilities predicted by multiple classifiers is most sensible when the constituent classifiers are appropriately calibrated. In our observation, both classifiers based on accelerometry and GNSS data are sufficiently calibrated. However, even if the PF algorithm does not yield an accurate estimate of the class probabilities, it makes the correct classification decision as long as the target class is more probable than the other ones. This holds even when the probability estimates are grossly inaccurate, which is inconsequential in many applications. Therefore, the underlying approximate probability model of the PF algorithm does not necessarily hurt its robustness.

A minor disadvantage of the PF algorithm with respect to the FC algorithm is that it requires the prior occurrence probabilities of the classes. This does not necessarily pose any challenge or excessive burden as the prior class probabilities can be easily calculated for any dataset during training. It is important to ensure that this prior knowledge is appropriate for the real-world settings in which the animal behavior classification algorithm is deployed.

The computational complexity and memory requirement of the examined FC and PF algorithms are only moderately higher compared with those of the classification algorithm based on the accelerometry data alone, i.e. the algorithm proposed in~\citep{RA21}. Therefore, both FC and PF algorithms run seamlessly on the embedded systems of our collar and ear tags and fulfill in-situ classification of animal behavior.

Despite their usefulness and essential role in evaluating the performance of different algorithms considered in this work, our datasets have some limitations. First, the data collection took place in a rather small area of 25 x 25 m, albeit alternated daily. Such dimensions do not correspond to the true conditions that most grazing beef cattle experience during most of their lives. However, the confinement was required to make the behavior of the cattle visible through the cameras placed at the corners of the experiment area. This was in turn to enable accurate post-hoc annotations via inspecting the recorded footage as well as collecting high-quality video data for other research activities. As a result, the cattle did not require nor had much opportunity to walk within the paddock. Hence, most of our data labeled as walking relates to the occasions when the cattle were driven from the overnight cattle pen to the experiment paddock and back. This has led to a sizable proportion of walking instances to occur far from the water point. An unnatural consequence of this is that the DtWP feature carries significant information about the walking behavior compared to the speed feature. This might have hindered true evaluation of the use of GNSS speed data for improving the classification accuracy of walking behavior in grazing cattle.

Another detriment of the small experiment areas used for the data collection is that the average GNSS position estimation error of approximately seven meters is substantial relative to the area the cattle were contained within. This limits the utility of the DtWP feature. In our future field trials, we plan to employ differential or real-time kinematics GNSS positioning for enhanced precision and expand the experiment areas as much as practical.

Classifying rare or infrequent animal behaviors such as the drinking behavior of grazing cattle is fundamentally challenging. This is primarily due to the paucity of training data that can practically be obtained for such behaviors. Annotating the rare behaviors is also difficult since they happen occasionally and in short durations. However, multimodal animal behavior classification can help tackle this problem to some extent through using data of multiple sensing modes, which may contain rich information about some rare behavior classes.

The alia class encompasses all behaviors apart from grazing, walking, resting, and drinking. It includes many behaviors such as grooming, scratching, suckling, calling (vocalizing), defecating, and urinating, which involve various animal body poses and movement intensity levels. This makes the alia class hard to distinguish from the other better-specified behavior classes. The scarcity of training data for this class further adds to the challenge. In future work, we will explore solutions that may obviate the need for defining any such class.

\section{Conclusion}

We studied the classification of animal behavior via the joint use of relevant data available through multiple sensing modes, i.e., accelerometry and GNSS, using two approaches. One approach is based on concatenating the features associated with all sensing modes and the other is based on fusing the posterior probabilities of behavior classes inferred from the features associated with each sensing mode. Both approaches are computationally efficient and readily implementable on the embedded systems of the smart collar and ear tags, which we use for monitoring cattle and classifying their behavior in situ and in real time. Our cross-validated performance evaluations using two datasets collected from grazing cattle via our collar and ear tags revealed that both approaches result in improved classification accuracy compared with using the data of either sensing mode alone. However, the second approach, i.e., posterior probability fusion, led to superior classification accuracy and efficiency despite its reliance on the assumption that the features extracted from the data of different sensing modes are conditionally independent given the behavior class.

\section*{Acknowledgment} \label{sec:acknowledgement}

This research was undertaken with strategic investment funding from the CSIRO and NSW Department of Primary Industries. We would like to thank the technical staff who were involved in the research at CSIRO FD McMaster Laboratory Chiswick, i.e., Flavio Alvarenga, Alistair Donaldson, and Reg Woodgate of the NSW Department of Primary Industries, and Jody McNally and Troy Kalinowski of the CSIRO Agriculture and Food. We also recognize the contributions of the CSIRO Data61 staff who have designed and built the hardware and software of the devices used for data collection, specifically, Lachlan Currie, John Scolaro, Jordan Yates, Leslie Overs, and Stephen Brosnan.

\bibliographystyle{elsarticle-harv}
\bibliography{references}

\end{document}